\documentclass[journal]{IEEEtran}
\ifCLASSINFOpdf
\else
\fi
\usepackage{color}
\usepackage{caption}
\usepackage{float}
\usepackage{diagbox}
\usepackage{mathtools,amssymb}
\usepackage{lineno,hyperref}
\usepackage[numbers]{natbib}

\usepackage{algpseudocode}
\usepackage{subfigure}

\modulolinenumbers[5]
\usepackage{verbatim}
\usepackage{amsmath}
\usepackage{amsmath,amsfonts,bm}

\def\eqref#1{equation~\ref{#1}}
\def\1{\bm{1}}

\DeclareMathAlphabet{\mathsfit}{\encodingdefault}{\sfdefault}{m}{sl}
\SetMathAlphabet{\mathsfit}{bold}{\encodingdefault}{\sfdefault}{bx}{n}

\usepackage{multirow}
\usepackage{algorithm}
\usepackage{algorithmicx}
\usepackage{algpseudocode}
\usepackage{url}
\usepackage{graphicx}
\usepackage{amssymb}
\usepackage{wrapfig}
\usepackage{xcolor}
\newcommand{\revise}[1]{{\color{black}#1}}

\begin{document}
\title{A Class-Aware Representation Refinement Framework for Graph Classification}

\author{Jiaxing~Xu,
        Jinjie~Ni,
        and~Yiping~Ke% <-this % stops a space
\thanks{J. Xu, J. Ni and Y. Ke are with the School of Computer Science and Engineering, Nanyang Technological University, Singapore. E-mail: jiaxing003@e.ntu.edu.sg, jinjie001@e.ntu.edu.sg, ypke@ntu.edu.sg.}}

\maketitle

\begin{abstract}
Graph Neural Networks (GNNs) are widely used for graph representation learning. Despite its prevalence, GNN suffers from two drawbacks in the graph classification task, the neglect of graph-level relationships, and the generalization issue. Each graph is treated separately in GNN message passing/graph pooling, and existing methods to address overfitting operate on each individual graph. This makes the graph representations learnt less effective in the downstream classification. In this paper, we propose a Class-Aware Representation rEfinement (CARE) framework for the task of graph classification. CARE computes simple yet powerful class representations and injects them to steer the learning of graph representations towards better class separability. CARE is a plug-and-play framework that is highly flexible and able to incorporate arbitrary GNN backbones without significantly increasing the computational cost. We also theoretically prove that CARE has a better generalization upper bound than its GNN backbone through Vapnik-Chervonenkis (VC) dimension analysis. Our extensive experiments with 11 well-known GNN backbones on 9 benchmark datasets validate the superiority and effectiveness of CARE over its GNN counterparts.
\end{abstract}

% \begin{IEEEkeywords}
% Graph Nrueal Network, Representation Learning, j, \LaTeX, paper, template.
% \end{IEEEkeywords}

% \fi
\IEEEpeerreviewmaketitle

\section{Introduction}
In recent years, Graph Neural Networks (GNNs) have attracted a surge of attention and been extensively used to learn graph representations for downstream tasks such as graph classification \cite{xu2018powerful, ying2018hierarchical}, \revise{link prediction \cite{li2023imf}, graph clustering \cite{kipf2016variational}}, etc. They have been applied to graph-structured data from a wide range of application domains, including \revise{social networks \cite{kipf2016semi, hamilton2017inductive}, molecular data \cite{gilmer2017neural}, recommendation system \cite{bian2023cpmr}, brain networks \cite{10508252}}, and many more. 

The GNN-based graph representation learning can be divided into two categories: (1) \revise{some of the methods \cite{lee2019self}} use a down-sampling strategy \cite{wu2020comprehensive} to aggregate and transform node representations to graph representations; (2) while some other methods \cite{ying2018hierarchical, zhang2019hierarchical, nouranizadeh2021maximum} obtain the graph embedding by exploring the hierarchical structure of each input graph. These approaches suffer from two major drawbacks when applied to the downstream classification task: (1) Neglect of graph-level relationships; and (2) Generalization issue. 

% \noindent
\textbf{Neglect of graph-level relationships.} Existing GNN architectures consider each input graph independently in their training processes. Input graphs are passed individually to GNN to yield node representations. \revise{In addition, the model also treats each graph separately in its loss design.} The relationships (similarity and/or discrepancy) among different input graphs are fully neglected. Though the model parameters are trained by the set of input graphs collectively, it is done through a long pathway from node representations to graph representations and finally to the loss. As a result, the effectiveness of the model will be significantly compromised when applied to the downstream classification. With molecular data, for instance, one would want the molecules from the same class to share similar representations. \revise{This is natural as molecules that belong to the same class often carry certain common substructures} (e.g., the same set of functional groups). These substructures could be class-specific and serve as a perfect signature of a class. Without considering such graph-level information, \revise{the learnt graph representations would be less effective in separating different graph classes.}

% \noindent
\textbf{Generalization issue.} This is an inherent issue in GNN that the model tends to overfit when the network gets deeper or the hidden dimensionality gets larger \cite{song2021network}. Some methods have been proposed to alleviate this issue. Several graph-argumentation based methods improve the generalization ability by modifying input graphs \cite{papp2021dropgnn}, or generating new graphs for adversarial learning \cite{ding2018semi} and contrastive learning \cite{you2020graph}. Some other works \cite{byrd2019effect, lin2017focal} resample or reweight data instances to remit the overfitting problem. However, these methods operate on each individual graph and fail to explore the effectiveness of graph-level information in improving generalization. 

In this paper, \revise{we develop a Class-Aware Representation rEfinement (CARE) framework to address the two above-mentioned limitations of GNNs in graph classification tasks.} Built upon the two existing steps of GNN models for node representation learning and graph pooling, we introduce a new block for extracting class-specific information, namely the Class-Aware Refiner. The idea of this refiner is simple but effective. Under the supervision of ground-truth labels, the refiner learns the class representation from a bag of subgraph representations (generated by graph pooling) from each class, and uses it to refine the representation of each graph within the same class. In this way, we inject the class-specific information to graph representations, with the hope that it can steer the graph representation learning to reflect class signatures. Inspired by the separability in clustering \cite{wen2016discriminative},  we also propose a class loss that takes into account intra-class graph similarity and inter-class graph discrepancy. This class loss is combined with the classification loss to directly influence the training of class representations to gain better class separability. We further design two different architectures of CARE so that our framework can incorporate arbitrary GNN backbones. \revise{Regarding model generalization, we analyze the Vapnik-Chervonenkis (VC) dimension \cite{vapnik2015uniform} of CARE and provide a theoretical guarantee that the upper bound of the VC dimension of CARE is lower than that of the GNN backbone.} The better ability of CARE in alleviating overfitting is also evidenced in our experiments.

We empirically validate the graph classification performance of CARE on 9 benchmark datasets with 11 commonly used GNN backbones. The results demonstrate that CARE significantly outperforms its GNN backbones: it achieves up to 11\% improvement in classification accuracy. We also perform a series of ablation studies to assess the effect of each component. We use a case study to showcase that the graph representations refined by CARE is able to achieve better class separability. Though CARE introduces an additional refiner to the backbone model, our results show that the consideration of graph-level information drives the model to converge with fewer training epochs. Consequently, the training time of CARE is comparable to or even shorter than its GNN counterparts. To sum up, by seamlessly integrating CARE into a GNN model, it can effectively incorporate graph-level relationships, leading to performance improvements and enhanced generalization ability without augmenting the time and memory complexity of the base model.

The main contributions of this paper are summarized as follows:

\begin{itemize}
\item We propose a novel graph representation refinement framework CARE, which considers class-aware graph-level relationships. CARE is a flexible plug-and-play framework that can incorporate arbitrary GNNs without significantly increasing the computational cost. 
\item We provide theoretic support through VC dimension analysis that CARE has better generalization upper bound in comparison with its GNN backbone.
\item We perform extensive experiments using 11 GNN backbones on 9 benchmark datasets to justify the superiority of CARE on graph classification tasks in terms of both effectiveness and efficiency.
\end{itemize}

\section{Related Works}
\label{app:related_works}

\subsection{Graph Neural Networks}

Kipf and Welling \cite{kipf2016semi} were the first to introduce the convolution operation to graphs. Later on, Hamilton et al. \cite{hamilton2017inductive} proposed to use sampling and aggregation to learn node representations. The attention mechanism was introduced to graph convolution in \cite{velivckovic2017graph} to yield node representations by unequally considering messages from different neighbors. The Graph Isomorphism Network (GIN) was proposed in \cite{xu2018powerful}, which has been shown to be as powerful as the Weisfeiler-Lehman (WL) test \cite{weisfeiler1968reduction}. Recently, the transformer architecture \cite{vaswani2017attention} has been introduced to GNN \cite{nguyen2019universal, rong2020grover}, which considers the relatedness between nodes in the node representation learning. 

In addition to exploring novel graph aggregation methods, several existing works have proposed versatile extensions that integrate GNNs to further enhance the performance of base models. For instance, GIB \cite{yu2020graph} presents a novel objective focused on recognizing maximally informative subgraphs, employing a bi-level optimization scheme and connectivity loss to optimize the GIB objective. \revise{Huang et al. \cite{huang2022graph} introduce a random filter to optimize the graph convolution layer, aiming to accelerate training on large graphs without compromising effectiveness.} 

For GNN-based graph classification, graph pooling is commonly applied as the readout function to generate graph representations. Existing pooling methods can be categorised under \revise{node drop pooling \cite{ zhang2018end, lee2019self} and node clustering pooling \cite{ying2018hierarchical, zhang2019hierarchical, nouranizadeh2021maximum}}. Node drop pooling uses learnable scoring functions to drop nodes with lower scores while node clustering pooling casts the graph pooling problem into the node clustering problem \cite{baek2021accurate}. 
In a recent development, Yang et al. \cite{yang2021soft} leverage the mask mechanism to select the subgraph by considering the consistency over samples. 
\revise{Both categories of the pooling methods focus on node-level manipulations and neglect the graph-level information.}

\subsection{Methods to Treat GNN Overfitting}

Expressiveness of GNN could be improved by increasing the number of model parameters through e.g., expanding the hidden dimension of the GNN layer or adding more layers. However, this process could detriment the performance and induce overfitting \cite{song2021network}.

A typical method to treat GNN overfitting is via data augmentation. Several works \cite{papp2021dropgnn, rong2019dropedge} used node/edge dropping augmentations. \cite{ding2018semi} proposed a generator-classifier network under the adversarial learning setting to generate fake nodes. \cite{feng2019graph} performed adversarial perturbations to node features. Recently, the paradigm of contrastive learning has been introduced to GNN to perform graph contrastive learning \cite{you2020graph, thakoor2021bootstrapped}. SUGAR \cite{sun2021sugar} generates subgraphs and uses these subgraphs for reconstruction, which still follows the graph contrastive learning paradigm. All these methods perform augmentation on individual graphs and again neglect the graph-level information.

Several resampling \cite{byrd2019effect, zhou2020k} and reweighting \cite{lin2017focal, shi2021boosting} methods have also been proposed to prevent GNN from overfitting. In general, these methods design algorithms to control the influence of each sample on the model, while the sample relations are not taken into account.

To the best of our knowledge, our work is the first to explore the use of class-aware graph-level relationships to alleviate the overfitting in GNNs.

\section{Our Proposed Method}
\label{sec:method}

In this section, we first give the problem formulation of graph classification, \revise{followed by the description of our proposed CARE framework.} We also discuss two architectures for applying CARE to existing GNNs. Finally, \revise{we provide theoretical support for the generalization of CARE.}

\subsection{Problem Formulation}
\label{subsec:problem}
We represent a graph as $G = (\boldsymbol{A}, \boldsymbol{X})$, where $\boldsymbol{A} \in \{0, 1\}^{n \times n}$ is its adjacency matrix, and $\boldsymbol{X} \in \mathbb{R}^{n \times c}$ denotes the feature matrix with each node characterized by a feature vector of $c$ dimensions. The node set of $G$ is denoted by $\mathcal{V}_G$ and $|\mathcal{V}_G| = n$. We use $\boldsymbol{X}_{v}$ to denote the feature vector of a node $v \in \mathcal{V}_G$. Table \ref{tab:notation} summarizes the notations used throughout the paper.

\begin{table}[h]
\centering
\small
\caption{Notation Table}
\begin{tabular}{cc}
\hline
Notation                         & Description                       \\ \hline
$G$                              & An input graph                             \\
$\boldsymbol{A}$                 & Adjacency matrix of $G$                 \\
$\boldsymbol{X}$                 & Node feature matrix of $G$                     \\
$\mathcal{V}_G$                  & Node set of $G$             \\
$v$                              & A node in G                              \\
$n$                              & Number of nodes in $G$          \\
$c$               & Dimensionality of node feature vector $\boldsymbol{X_v}$\\
$\mathcal{D}$                    & Input dataset                          \\
$\mathcal{G}$                    & Input graph set                         \\
$\mathcal{Y}$                    & Input label set                         \\
$\mathcal{B}_i$                & Bag of (sub)graph representations of class $i$ \\
$y_G$                            & Label of $G$                \\
$\boldsymbol{H}$                 & Node representations              \\
$l$                              & Number of layers in GNN    \\
$\boldsymbol{hg}_G$                           & Whole graph representation of $G$       \\
$\boldsymbol{H}_v$                & Node representation of $v$         \\
$m$        & Dimensionality of node representation $\boldsymbol{H}_v$\\
$\boldsymbol{G}_{sub}$           & Subgraph of $G$             \\
$\boldsymbol{A}_{sub}$           & Adjacency matrix of $G_{sub}$     \\
$\boldsymbol{H}_{sub}$           & Node representations of $G_{sub}$ \\
$i$                              & $i$-th class in $\mathcal{Y}$               \\
$\boldsymbol{hc}_i$                           & Class representation of class $i$   \\
$\boldsymbol{hg}_G^{\prime}$                  & Refined representation of $G$ \\
$\hat{y}$                        & Predicted class label                   \\ \hline
\end{tabular}
\label{tab:notation}
\end{table}

Given a data set of labeled graphs $\mathcal{D} = (\mathcal{G}, \mathcal{Y}) = \{(G, y_G)\}$, where $y_G \in \mathcal{Y}$ is the corresponding label of graph $G \in \mathcal{G}$, the problem of graph classification aims to learn a predictive function $f$: $\mathcal{G} \rightarrow \mathcal{Y}$ that maps graphs to their labels.

\subsection{Class-Aware Representation Refinement Framework}
\label{subsec:care}

\begin{figure*}[h]
  \centering
  \includegraphics[width=.9\linewidth]{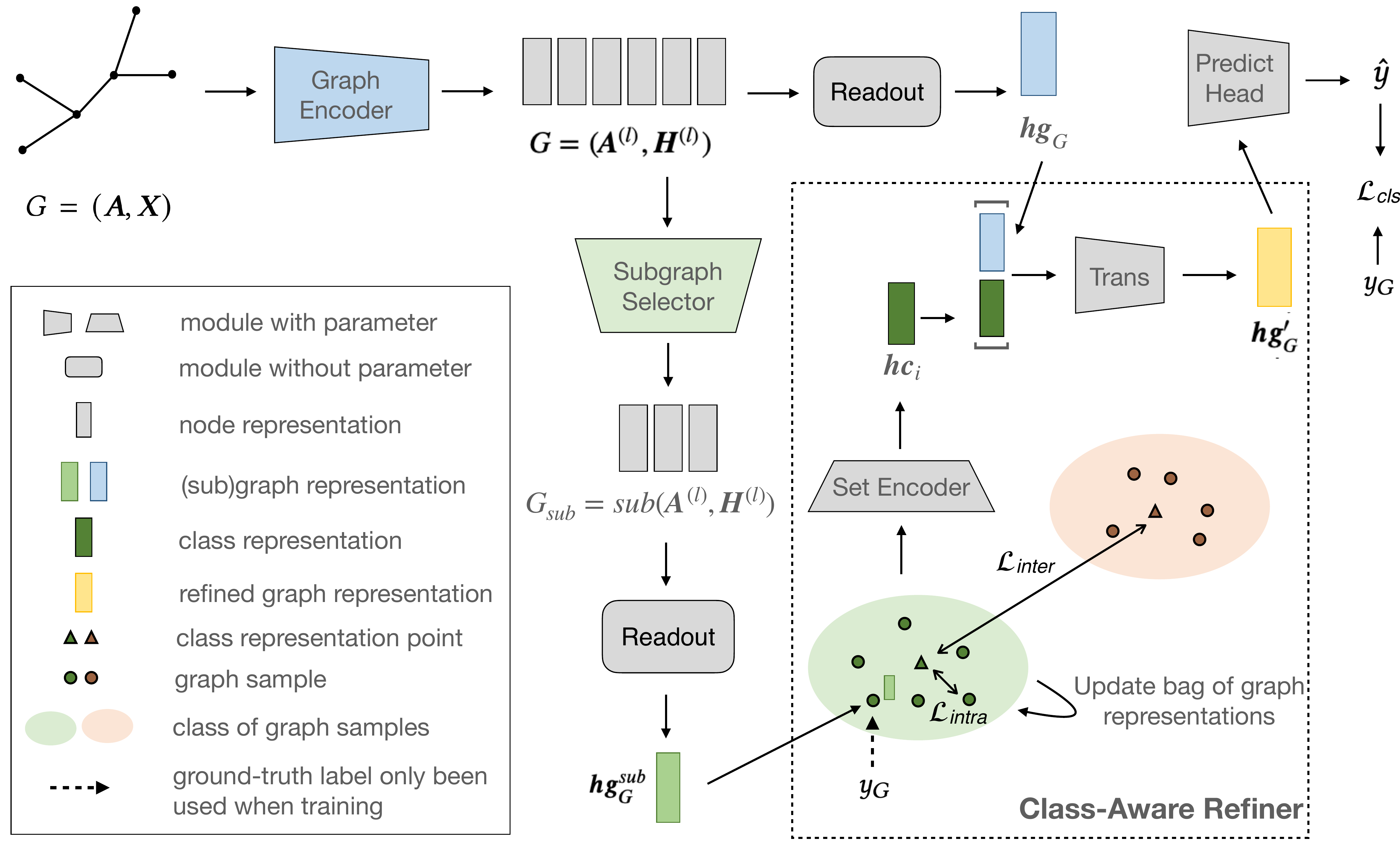}
  \caption{Framework of CARE. CARE contains four main components, including a graph encoder, a subgraph selector, a class-aware refiner and a class loss. The former two allow the flexible incorporation of existing GNN methods, while the latter two are newly proposed in our framework. We introduce them sequentially in Section \ref{subsec:care}.}
  \label{fig:framework}
\end{figure*}

We now describe our proposed Class-Aware Representation rEfinement framework (CARE), which aims to refine graph representations by considering the graph-level similarity. CARE contains four main components, including a graph encoder, a subgraph selector, a class-aware refiner and a class loss. The former two allow the flexible incorporation of existing GNN methods, while the latter two are newly proposed in our framework. The learned class representation plays a pivotal role in incorporating class-specific information to refine the graph representation, thereby addressing the limitation of neglecting the graph-level relationship. Additionally, the class loss serves to underscore the similarity within the same class and the disparity between different classes. By highlighting class signatures and mitigating the focus on individual outliers, it effectively alleviates generalization issues. Figure \ref{fig:framework} depicts the CARE framework. 

The remainder of this section describes the four components in detail. We first introduce the graph encoder to get the initial node/graph representation, and then describe the subgraph selector to extract an appropriate substructure for the subsequent class representation learning. The class-aware refiner learns class representations from different graph classes, which are used to refine graph representations. A class loss is proposed to further improve class separability. The two new components in CARE only contain a small number of parameters and are easy to plugin arbitrary GNN backbone. 

\noindent
\textbf{Graph Encoder.} A graph encoder extracts the node representations $\boldsymbol{H}$ and the graph-level representation $\boldsymbol{hg}_G$ for graph $G$. CARE does not impose any constraint on the architecture of the graph encoder. Any message-passing GNN model could be applied here, which is formalized as 
\begin{equation}
\boldsymbol{H}_{v}^{(l+1)}=\operatorname{UPDATE}^{(l)}\left(\boldsymbol{H}_{v}^{(l)}, \operatorname{ AGG }^{(l)}\left(\left\{\boldsymbol{H}_{u}^{(l)}\right\}_{\forall u \in \mathcal{N}(v)}\right)\right),
\label{eq:message_passing}
\end{equation}
where $\boldsymbol{H}^{(l+1)} \in \mathbb{R}^{n \times m}$ denotes the $(l+1)$-th layer node representation with $m$ dimensions, $\operatorname{UPDATE}$ and $\operatorname{AGG}$ are arbitrary differentiable update and aggregate functions, $\mathcal{N}(v)$ represents the neighbor node set of node $v \in \mathcal{V}_G$, and $\boldsymbol{H}^{(0)}_v$ is initialized as the input feature vector $\boldsymbol{X}_v$.

After a few message-passing layers, we can obtain a set of node representations. A $\operatorname{READOUT}$ function can be applied to produce the graph representation $\boldsymbol{hg}_{G} \in \mathbb{R}^{m}$ as $\boldsymbol{hg}_{G}=\operatorname{READOUT}\left(\left\{\boldsymbol{H}_{v} \mid v \in \mathcal{V}_G\right\}\right)$.

\noindent
\textbf{Subgraph Selector.} The class-aware refiner in CARE aims to maintain generic features for graph samples from different classes. However, the READOUT function treats all nodes equally without considering the class information. In fact, \revise{graphs belong to different classes are likely to have various substructures.} To address this limitation, CARE introduces a subgraph selector $\operatorname{sub}(\cdot)$ to filter nodes in the original graph, which is defined as:

\begin{equation}
\boldsymbol{A}^{(l+1)}, \boldsymbol{H}^{(l+1)} = \operatorname{sub}(\boldsymbol{A}^{(l)}, \boldsymbol{H}^{(l)}).
\label{eq:pooling}
\end{equation}

Any graph pooling methods could be applied here to select subgraphs. Typical ones include node drop pooling methods \cite{lee2019self} and node clustering pooling methods \cite{baek2021accurate}. Besides, a similar challenge of finding the informative substructure from the whole picture arises in the task of point cloud registration within the computer vision domain, termed incomplete overlap estimation. \revise{Recent efforts have tackled this challenge by focusing on extracting reliable overlapping representations before registration \cite{wu2023rornet}.}

\noindent
\textbf{Class-Aware Refiner.} As existing GNN models ignore the relationships of graphs from different classes, a new component is designed in CARE to fill this gap. In the training process, the class-aware refiner utilizes the ground truth label of each training instance. It maintains a bag of encoded (sub)graph representations $\mathcal{B}_i$ and aggregates these representations to obtain a class representation $\boldsymbol{hc}_i$ for each class $i \in \mathcal{Y}$. The aggregation function is a universal Set Encoder, e.g., DeepSets \cite{zaheer2017deep} or PointNet \cite{CharlesRQi2016PointNetDL}. Herein, we apply DeepSets as in Eq. (\ref{eq:deepsets}), in which $\rho(\cdot)$ is a multilayer perceptron (MLP) with a non-linear function ReLU, and $\phi(\boldsymbol{hg}) = \boldsymbol{hg}/{|\mathcal{B}_i|}$. 

\begin{equation}
\boldsymbol{hc}_i=\rho\left(\sum_{\boldsymbol{hg}_G^{sub} \in \mathcal{B}_i} \phi(\boldsymbol{hg}_G^{sub})\right) ,
\label{eq:deepsets}
\end{equation}

\noindent
where $\boldsymbol{hg}_G^{sub}$ is the subgraph representation of a graph $G \in \mathcal{G}$, obtained by passing the output of the subgraph selector through the READOUT function. 

In general, given a training sample $G$ belongs to class $i$, it is first encoded by the graph encoder. A subgraph selector is then applied to generate a subgraph representation for this graph. Such subgraph representation is used to update the class representation of its corresponding class by Eq.(\ref{eq:deepsets}). The class representation $\boldsymbol{hc}_i$ is used to refine the graph representation for graph $G$:

\begin{equation}
\boldsymbol{hg}_G^{\prime} = \operatorname{Trans}([\boldsymbol{hg}_G \mid \boldsymbol{hc}_i]),
\label{eq:refine}
\end{equation}

\noindent
where $\operatorname{Trans}(\cdot)$ is a transformation function and $\boldsymbol{hg}_G^{\prime}$ is the refined graph representation. $\mid$ denotes the concatenation operation that combines the original graph representation and its corresponding class representation into a unified representation. Herein, we also apply an MLP with a non-linear function ReLU as the transformation function $\operatorname{Trans}(\cdot)$. The training algorithm of the class-aware refiner is summarized in Algorithm \ref{alg:care}.

\begin{algorithm}[h]
\caption{Training of Class-Aware Refiner}
\label{alg:care}
\small
\noindent
\begin{flushleft}
\textbf{Input}: Subraph representation $\boldsymbol{hg}_G^{sub}$ of graph $G$ (output by Subgraph Selector), graph representation $\boldsymbol{hg}_G$, and ground-truth label $y_G$ of $G$; \\
\noindent
\textbf{Output}: Refined graph representation $\boldsymbol{hg}_G^{\prime}$, intra-class loss $\mathcal{L}_{intra}$ and inter-class loss $\mathcal{L}_{inter}$;
\end{flushleft}
\begin{algorithmic}[1]
\Statex Use $\boldsymbol{hg}_G^{sub}$ to update the bag of (sub)graph representations $\mathcal{B}_i$ for $i = y_G$;
\Statex Calculate the class representation $\boldsymbol{hc}_{i}$ by Eq. (\ref{eq:deepsets});
\Statex Use $\boldsymbol{hc}_{i}$ and $\boldsymbol{hg}_G$ to obtain the refined graph representation $\boldsymbol{hg}_G^{\prime}$ by Eq. (\ref{eq:refine});
\Statex Calculate the intra-class loss $\mathcal{L}_{intra}$ and inter-class loss $\mathcal{L}_{inter}$ by Eqs. (\ref{eq:pos_loss}) and (\ref{eq:neg_loss});
\end{algorithmic}
\end{algorithm}

In the validation and the test processes, the ground truth label $y_G$ \revise{is unavailable for a validation/test graph $G$.} \revise{However, we need a label for each validation/test graph to determine the appropriate class representation for composing its refined representation.} In this case, the Class-Aware Refiner will predict a pseudo label $\widetilde{y}_G$ for graph $G$ by classifying it to the most similar class and use the corresponding class representation for graph representation refinement. We use the cosine similarity as a metric to quantify the similarity between a graph representation and a class representation. The pseudo label is obtained as $\widetilde{y}_G = \underset{i \in \mathcal{Y}}{\arg\max}({\operatorname{cos\_sim}(\boldsymbol{hg}_G^{sub}, \boldsymbol{hc}_i)})$. Note that class representations are kept unchanged in the validation/test process.

\noindent
\textbf{Class Loss.} \revise{For the graph classification task}, it would be beneficial to exploit the graph similarity within the same class and the graph discrepancy between different classes. This is essential for making different classes more separable. Using the classification loss only fails to learn such graph-level relations. Therefore, a class loss $\mathcal{L}_{class}(\cdot)$ is proposed in CARE to enforce the intra-class similarity and the inter-class discrepancy. The former $\mathcal{L}_{intra}$ is defined as the similarity between each graph representation and its class representation, while the latter $\mathcal{L}_{inter}$ is defined as the similarity between different class representations:

\begin{equation}
\mathcal{L}_{intra} = \underset{i \in \mathcal{Y} }{AVG}(\underset{y_G=i}{AVG}(\operatorname{cos\_sim}(\boldsymbol{hc}_i, \boldsymbol{hg}_G^{sub}))),
\label{eq:pos_loss}
\end{equation}

\begin{equation}
\mathcal{L}_{inter} = \underset{i \in \mathcal{Y}}{AVG}(\underset{j > i, j \in \mathcal{Y}}{AVG}(\operatorname{cos\_sim}(\boldsymbol{hc}_i, \boldsymbol{hc}_j))),
\label{eq:neg_loss}
\end{equation}

We again use the cosine similarity as a metric. The class loss $\mathcal{L}_{class} = \exp (\mathcal{L}_{inter} - \lambda_1 * \mathcal{L}_{intra})$ is then defined as a function that maximizes $\mathcal{L}_{intra}$ and minimizes $\mathcal{L}_{inter}$, where $AVG(\cdot)$ is the average function and $\lambda_1$ is a trade-off hyperparameter.

The predicted class label is still supervised by a classification loss $\mathcal{L}_{cls}(\cdot)$. Herein, we apply the commonly used cross-entropy loss \cite{cox1958regression}. The overall loss $\mathcal{L}$ of CARE is defined as: 

\begin{equation}
\mathcal{L} = \mathcal{L}_{cls} + \lambda_2 * \mathcal{L}_{class},
\label{eq:total_loss}
\end{equation}

\noindent
where $\lambda_2$ is a trade-off hyperparameter for balancing the classification loss and the class loss.

\subsection{Model Architecture Variants}
\label{subsec:model_arch}

\revise{Given that GNNs can be categorised into hierarchical and non-hierarchical types, we have designed two corresponding architectures that integrate CARE as a plug-and-play module for different GNN backbones.}

\noindent
\textbf{Global Architecture.} Several GNN models (e.g., GCN \cite{kipf2016semi}, GAT \cite{velivckovic2017graph} and GraphSAGE \cite{hamilton2017inductive}) apply the readout function only at the end of graph convolution. The global architecture of CARE is designed to apply the Class-Aware Refiner only after the readout function. The outputs are then passed to a linear layer for graph classification.

\noindent
\textbf{Hierarchical Architecture.} Some other GNN models, such as GIN \cite{xu2018powerful} and UGformer \cite{nguyen2019universal}, have a readout function in each graph convolutional layer. The graph representations from each layer are taken into account when making the final prediction. The hierarchical architecture of CARE is designed to apply the class-aware refiner on each layer in order to cope with the hierarchical GNN backbones.

\subsection{Generalization Analysis}
\label{subsec:generalization}

In this section, we present the theoretical support that the proposed CARE has a better model generalization than its GNN backbone in the case of binary classification. We use the VC dimension to measure the capacity of a model. Based on the VC theory \cite{vapnik2000nature}, reducing the VC dimension of a model has the effect of eliminating potential generalization errors. 

Our analysis is grounded on the VC theory for neural nets \cite{bartlett2003vapnik}: the VC dimension of a neural network is upper bounded by a function with respect to the number of model parameters $t$ and the number of operations $p$. In the following, we first derive the computational complexity of the GNN backbone and CARE measured by the number of multiplications, based on which we obtain an upper bound of the VC dimension for each model. We then present a theorem that states that CARE has a lower VC dimension upper bound than its GNN backbone when the number of parameters is identical. In subsequent discussions, we use GCN as an example backbone. The conclusion generally applies to other backbones by plugging in their corresponding computational complexity. \revise{We present the theoretical results here and defer the detailed proofs to Appendix \ref{app:proofs}.}

\noindent
{\bf Computational Complexity of Models.} CARE and its backbone GCN are both composed of GCN layers, an embedding layer, and several fully-connected layers. 

[{\it Complexity of GCN Backbone}.] We denote the GCN layer in the GCN model as $\operatorname{gcn}(\cdot)$ and its input/output dimensions as $h_{gcn_{in}}$/$h_{gcn_{out}}$. The layer mapping is given by $\operatorname{gcn}(\boldsymbol{A}, \boldsymbol{H}) = \sigma_{gcn} (\boldsymbol{\hat{A}} \boldsymbol{H} \boldsymbol{W}_{gcn})$, where $\sigma_{gcn}$ is the activation function, $\boldsymbol{W}_{gcn} \in \mathbb{R}^{h_{gcn_{in}} \times h_{gcn_{out}}}$ is the weight matrix, and $\boldsymbol{\hat{A}}$ is the normalized adjacency matrix. The computational complexity of the GCN network measured by the multiplication number, denoted as $q_1(d)$, for $d$ number of layers, is given by:

\begin{equation}
\label{eq:q1}
    q_1(d) = \sum_{l=0}^d (n^2 h_{gcn_{in}}^l + n h_{gcn_{in}}^l h_{gcn_{out}}^l).
\end{equation}

[{\it Complexity of CARE}.] The GCN-based CARE network with a hierarchical architecture is composed of the following:

\begin{itemize}
\item a GCN layer same as the GCN backbone, whose computational complexity is $q_{gcn}^l = n^2 h_{gcn_{in}}^l + n h_{gcn_{in}}^l h_{gcn_{out}}^l$.
\item a subgraph selector (SAGPool), \revise{which contains a scoring layer (GCN with 1-dimensional output) and a top-k pooling algorithm.} The complexity is $q_{subgraph}^l = n^2 h_{gcn_{out}}^l + n h_{gcn_{out}}^l$.
\item a class-aware refiner contains a set encoder Eq. (\ref{eq:deepsets}) and a transformation layer Eq. (\ref{eq:refine}). The mapping of the fully-connection layer $\operatorname{fc}(\cdot)$ from input $\boldsymbol{H}$ is given by $\operatorname{fc}(\boldsymbol{H}) = \sigma_{fc}(\boldsymbol{H} \boldsymbol{W}_{fc})$, where $\sigma_{fc}$ is the activation function, $\boldsymbol{W}_{fc} \in \mathbb{R}^{h_{fc_{in}}\times h_{fc_{out}}}$, and $h_{fc_{in}}/h_{fc_{out}}$ are the input/output dimensions. 
The complexities for the set encoder and the transformation layer are $q_{set}^l = nh_{set_{in}}^lh_{set_{out}}^l$ and $q_{trans}^l = nh_{trans_{in}}^lh_{trans_{out}}^l$, respectively.
\end{itemize}

Therefore, the computational complexity of the GCN-based CARE is given by:

\begin{equation}
\label{eq:q2}
\begin{split}
    q_2(d) = \sum_{l=0}^d (q_{gcn}^l + q_{subgraph}^l + q_{set}^l + q_{trans}^l).
\end{split}
\end{equation}

\noindent
{\bf VC Dimension Upper Bound.} Inspired by the theoretical analysis in \cite{kabkab2016size} that derives an upper bound of the VC dimension for a CNN model, we extend its result to a GCN model, as given by the following lemma. 

{\bf Lemma 1} Let $\mathcal{C}^d$ be the set of GCN models with $d$ convolutional layers. Let $\mathcal{H}^d \triangleq \{h_c : I \to \{0, 1\} | c \in \mathcal{C}^d\}$ be the set of boolean functions implementable by all GCNs in $\mathcal{C}^d$. The VC dimension of GCNs, as well as CARE, satisfies $\operatorname{VC}_{dim}(\mathcal{H}^d) \leqslant \alpha (d \cdot q(d))^2$ for some constant $\alpha$. Here, $q(d)$ is the computational complexity of the model under consideration. 

\noindent
{\bf VC Dimension Comparison.} We now compare the upper bounds of VC dimension on the GCN backbone and CARE, which is formalized by the following theorem.

{\bf Theorem 1.} Assume that the number of parameters in a GCN backbone and CARE is identical. Let {\it upperVC(GCN)} and {\it upperVC(CARE)} be the upper bounds of VC dimension on the two models, respectively, which are given by Lemma 1.  We have ${\it upperVC(GCN)} > {\it upperVC(CARE)}$.

Based on Theorem 1 and VC theory, we conclude that our CARE model exhibits a lower upper bound for the VC dimension compared to its GCN backbone. Consequently, under the condition of an identical number of parameters, GCN augmented with CARE possesses better generalization potential than the original GCN backbone. \revise{This theoretical insight suggests that CARE effectively mitigates the generalization issue of GNNs by increasing their upper bounds of the VC dimension.}

\section{Experimental Setup}
\label{app:setup}
In this section, we present the experimental setup, including datasets, GNN backbones and detailed model implementation.

\subsection{Datasets}

\begin{table}[h]
\centering
\caption{Statistics of Datasets.}
\begin{tabular}{lcccc}
\hline
\textbf{Dataset} &
  \multicolumn{1}{l}{\textbf{Graph\#}} &
  \multicolumn{1}{l}{\textbf{Class\#}} &
  \multicolumn{1}{l}{\textbf{Avg Node\#}} &
  \multicolumn{1}{l}{\textbf{Avg Edge\#}} \\ \hline
D\&D         & 1178 & 2 & 284.32 & 715.66 \\
PROTEINS     & 1113 & 2 & 39.06  & 72.82  \\
MUTAG        & 188  & 2 & 17.93  & 19.79  \\
NCI1         & 4110 & 2 & 29.87  & 32.30  \\
NCI109       & 4127 & 2 & 29.68  & 32.13  \\
FRANKENSTEIN & 4337 & 2 & 16.90  & 17.88  \\
Tox21        & 8169 & 2 & 18.09  & 18.50   \\ 
ENZYMES      & 600  & 6 & 32.63  & 62.14   \\  
OGBG-MOLHIV & 41127 & 2 & 25.50  & 27.50 \\ \hline
\end{tabular}
\label{tab:dataset}
\end{table}

Nine commonly used benchmark datasets were tested in our experiments. Eight of them were selected from TUDataset \cite{KKMMN2016} and include DD, PROTEINS, MUTAG, NCI1, NCI109, FRANKENSTEIN (FRANK), Tox21 and ENZYMES. The last dataset OGBG-MOLHIV was selected from Open Graph Benchmark \cite{hu2020open} and consists of 41K+ graphs. The statistics of the datasets are summarized in Table \ref{tab:dataset}.

\subsection{GNN Backbones}

We test the effectiveness of CARE on a wide range of GNN backbones, including GCN \cite{kipf2016semi}, GraphSAGE \cite{hamilton2017inductive}, GIN \cite{xu2018powerful}, GAT \cite{velivckovic2017graph}, GraphSNN \cite{wijesinghe2021new}, UGformer \cite{nguyen2019universal}, SAGPool \cite{lee2019self}, DiffPool \cite{ying2018hierarchical}, HGPSLPool \cite{zhang2019hierarchical}, GXN \cite{li2020graph}and MEWISPool \cite{nouranizadeh2021maximum}. We apply CARE on each of them and compare the performance with the original backbone model. Among the 11 models selected, GIN and UGformer are hierarchical ones. We thus apply the hierarchical architecture CARE on them. The global architecture CARE is applied to the rest models. A brief introduction of each model is provided as follow: 

\subsection{Implementation Details}
The default number of graph convolutional layers in both CARE and GNN backbones is 4. We use SAGPool with a pooling ratio of 0.5 as the default subgraph selector in CARE. Notice that we did not apply any subgraph selector on GNNs that are already equipped with their own pooling methods for substructure extraction. This includes SAGPool, DiffPool, HGPSLPool and MEWISPool. The trade-off hyperparameters $\lambda_1$ and $\lambda_2$ in Eq. (\ref{eq:total_loss}) are set to 1 by default. The whole network is trained in an end-to-end manner using \revise{the Adam optimizer}. We use the early stopping criterion, i.e., we stop the training once there is no further improvement on the validation loss during 25 epochs. The learning rate is initialized to $10^{-4}$ and the maximum number of epochs is set to 1000. We set the hidden size to 146 and batch size to 20 for all models. The only exception is DiffPool when tested on the D\&D dataset. Since the D\&D dataset has a large number of nodes (see Table \ref{tab:dataset}), the hidden size and batch size 
are set to 32 and 6 to achieve an acceptable number of parameters in DiffPool.

For TUdataset, we split it into 8:1:1 for training, validation and test. For all experiments of CARE and GNN backbones, we evaluate each model with the same random seed for 10-fold cross-validation. We use the scaffold splits for the OGBG-MOLHIV dataset and report the average ROC-AUC with 10 random seeds.
All the codes were implemented using \revise{PyTorch and Deep Graph Library packages}. The experiments were conducted in a Linux server with Intel(R) Core(TM) i9-10940X CPU (3.30GHz), GeForce GTX 3090 GPU, and 125GB RAM.

\section{Experiments}

In this section, we first assess the performance of CARE in comparison with state-of-the-art GNN backbones in Subsection \ref{subsec:main}. We then conduct ablation studies in Subsection \ref{subsec:ablation} to analyze the effects of different components in CARE. A case study is presented in Subsection \ref{subsec:case_study} to further investigate how CARE affects the separability of graphs from different classes. The sensitivity test of hyperparameters in CARE is discussed in Subsection \ref{subsec:hyperparameter}. Finally, an evaluation of model efficiency is performed in subsection \ref{subsec:time_efficiency}.

\textbf{GCN} \cite{kipf2016semi} is a mean pooling baseline with graph convolution network as a message-passing layer. 

\textbf{GraphSAGE} \cite{hamilton2017inductive} is a mean pooling baseline, which adopts sampling to obtain a fixed number of neighbors for each node. 

\textbf{GIN} \cite{xu2018powerful} is a sum pooling baseline that uses a learnable parameter to adjust the weight of the central node, thus improving the message-passing network's ability to distinguish different graph structures.  

\textbf{GAT} \cite{velivckovic2017graph} is a mean pooling baseline, which adopts an attention mechanism to learn the relative weights between the node and its neighbors. 

\textbf{GraphSNN} \cite{wijesinghe2021new} is a mean pooling baseline that brings the information of overlap subgraphs into the message passing scheme as a structural coefficient.

\textbf{UGformer} \cite{nguyen2019universal} is a sum pooling baseline which identifies meta-paths to transform the graph structure for node representation learning and adopts Transformer for aggregation.

\textbf{SAGPool} \cite{lee2019self} is a graph pooling method that uses graph convolution in graph pooling to consider both node features and graph structure. 

\textbf{DiffPool} \cite{ying2018hierarchical} is an end-to-end trainable graph pooling method that can produce hierarchical representations for graphs. 

\textbf{HGPSLPool} \cite{zhang2019hierarchical} is a pooling method that introduces a structure learning mechanism to refine graph structure after pooling. 

\textbf{GXN} \cite{li2020graph} is a sort pooling \cite{zhang2018end} baseline, which uses vertex infomax pooling to select nodes that can maximally express their corresponding neighborhoods. 

\textbf{MEWISPool} \cite{nouranizadeh2021maximum} is a pooling method, which introduces Shannon capacity to maximize the mutual information between the input graph and the output graph. 

\begin{table*}[h]
\setlength\tabcolsep{2.5pt}
\scriptsize
\centering
\caption{Graph Classification Results (Average Accuracy ± Standard Deviation). Winner in each backbone/dataset pair is highlighted in bold.}
\begin{tabular}{lccccccccc}
\hline
\multicolumn{2}{c}{Model}             & D\&D         & PROTEINS     & MUTAG         & NCI1         & NCI109       & FRANK & Tox21        & ENZYMES      \\ \hline
\multirow{2}{*}{GCN}       & Original & 71.02 ± 3.17 & 73.89 ± 2.85 & 77.52 ± 10.81 & 78.80 ± 2.01 & 75.06 ± 2.50 & 55.58 ± 0.11 & 88.14 ± 0.29 & 62.17 ± 6.33 \\
                          & CARE     & \textbf{72.15} ± 3.88 & \textbf{75.01} ± 2.91 & \textbf{79.30} ± 11.81 & \textbf{79.66} ± 1.71 & \textbf{77.39} ± 2.34 & \textbf{59.67} ± 3.00 & textbf{90.59} ± 0.55 & \textbf{65.00} ± 5.63 \\ \hline
\multirow{2}{*}{GraphSAGE} & Original & 72.18 ± 2.93 & 74.87 ± 3.38 & 75.48 ± 6.11  & 63.94 ± 2.53 & 65.46 ± 1.12 & 52.95 ± 4.01 & 88.36 ± 0.15 & 52.50 ± 5.69 \\
                          & CARE     & \textbf{73.26} ± 3.25 & \textbf{76.81} ± 3.30 & \textbf{81.97} ± 6.42  & \textbf{76.13} ± 2.03 & \textbf{75.87} ± 2.51 & \textbf{64.43} ± 3.15 & \textbf{90.14} ± 0.74 & \textbf{55.83} ± 6.88 \\ \hline
\multirow{2}{*}{GIN}       & Original & 73.10 ± 2.44 & 72.41 ± 4.45 & 89.36 ± 4.71  & 81.96 ± 2.03 & 81.01 ± 1.84 & 67.23 ± 1.93 & 92.10 ± 0.59 & 62.79 ± 7.64 \\
                          & CARE     & \textbf{76.32} ± 3.33 & \textbf{73.14} ± 3.45 & \textbf{90.47} ± 5.11  & \textbf{82.34} ± 2.11 & \textbf{82.15} ± 1.79 & \textbf{67.33} ± 2.74 & \textbf{92.43} ± 0.78 & \textbf{68.17} ± 7.05 \\ \hline
\multirow{2}{*}{GAT}       & Original & 74.25 ± 3.76 & 74.34 ± 2.09 & 77.56 ± 10.49 & 78.07 ± 1.94 & 74.34 ± 2.18 & \textbf{62.85} ± 1.90 & 90.35 ± 0.71 & 67.67 ± 3.74 \\
                          & CARE     & \textbf{75.55} ± 2.43 & \textbf{76.72} ± 1.74 & \textbf{79.33} ± 5.82  & \textbf{78.93} ± 1.69 & \textbf{76.71} ± 1.45 & 62.57 ± 2.37 & \textbf{90.76} ± 0.73 & \textbf{70.83} ± 5.54 \\ \hline
\multirow{2}{*}{GraphSNN}       & Original & 76.03 ± 2.59 & 71.78 ± 4.11 & 84.04 ± 4.09 & 70.87 ± 2.78 & 70.11 ± 1.86 & \textbf{67.17} ± 2.25 & 92.24 ± 0.59 & 67.67 ± 3.74 \\
                          & CARE     & \textbf{76.67} ± 1.52 & \textbf{74.02} ± 4.81 & \textbf{86.71} ± 7.31  & \textbf{72.25} ± 2.59 & \textbf{70.46} ± 2.90 & 66.87 ± 2.33 & \textbf{92.36} ± 0.58 & \textbf{68.17} ± 2.94 \\ \hline
\multirow{2}{*}{UGformer}  & Original & 75.51 ± 3.92 & 70.17 ± 5.42 & 75.66 ± 8.67  & 68.84 ± 1.54 & 66.37 ± 2.74 & 56.13 ± 2.51 & 88.06 ± 0.50 & 64.57 ± 4.53 \\
                          & CARE     & \textbf{76.23} ± 4.45 & \textbf{71.84} ± 3.87 & \textbf{77.66} ± 5.93  & \textbf{71.48} ± 2.25 & \textbf{66.92} ± 1.58 & \textbf{57.10} ± 2.27 & \textbf{88.21} ± 0.24 & \textbf{65.24} ± 5.91 \\ \hline
\multirow{2}{*}{SAGPool}   & Original & 71.46 ± 3.60 & 74.12 ± 3.46 & 78.12 ± 8.35  & 78.34 ± 1.96 & 76.15 ± 2.25 & 59.07 ± 2.23 & \textbf{90.78} ± 0.63 & 62.00 ± 4.76 \\
                          & CARE     & \textbf{73.28} ± 2.25 & \textbf{74.75} ± 3.14 & \textbf{79.81} ± 7.52  & \textbf{79.78} ± 1.67 & \textbf{76.44} ± 1.74 & \textbf{59.67} ± 2.04 & 90.64 ± 0.38 & \textbf{63.17} ± 4.37 \\ \hline
\multirow{2}{*}{DiffPool}  & Original & 70.45 ± 2.54 & 72.18 ± 2.80 & 85.26 ± 4.79  & 79.78 ± 2.11 & 76.98 ± 1.88 & 65.01 ± 3.17 & 91.02 ± 0.37 & 48.33 ± 6.67 \\
                          & CARE     & \textbf{72.90} ± 4.58 & \textbf{73.10} ± 3.94 & \textbf{89.00} ± 7.00  & \textbf{81.20} ± 2.27  & \textbf{80.43} ± 1.51 & \textbf{66.26} ± 2.11 & \textbf{91.61} ± 0.59 & \textbf{51.17} ± 6.75 \\ \hline
\multirow{2}{*}{HGPSLPool} & Original & 71.25 ± 3.25 & 73.06 ± 3.20 & 80.82 ± 6.63  & 79.26 ± 1.44 & 75.83 ± 1.98 & 60.82 ± 2.85 & 90.12 ± 0.47 & 63.33 ± 5.06 \\
                          & CARE     & \textbf{71.61} ± 3.36 & \textbf{75.47} ± 3.98 & \textbf{82.31} ± 6.91  & \textbf{79.77} ± 1.97 & \textbf{76.87} ± 1.94 & \textbf{63.36} ± 1.73 & \textbf{90.44} ± 0.69 & \textbf{66.00} ± 4.48 \\ \hline
\multirow{2}{*}{GXN}       & Original & 67.62 ± 5.85 & 70.32 ± 3.03 & 83.22 ± 7.97  & 73.34 ± 2.54 & 72.18 ± 2.24 & 60.86 ± 2.17 & 89.93 ± 0.73 & 63.13 ± 4.68 \\
                          & CARE     & \textbf{71.82} ± 4.30 & \textbf{72.70} ± 2.73 & \textbf{87.19} ± 6.61  & \textbf{74.75} ± 2.90 & \textbf{73.78} ± 1.66 & \textbf{62.64} ± 2.27 & \textbf{90.43} ± 0.76 & \textbf{64.44} ± 6.96 \\ \hline
\multirow{2}{*}{MEWISPool} & Original & \textbf{76.03} ± 2.59 & 68.10 ± 3.97 & 84.73 ± 4.73  & 74.21 ± 3.26 & 75.30 ± 1.45 & 64.63 ± 2.83 & 88.13 ± 0.05 & 53.66 ± 6.07    \\
                          & CARE     & 75.72 ± 2.54 & \textbf{69.64} ± 3.69 & \textbf{86.70} ± 4.27  & \textbf{76.48} ± 2.74 & \textbf{75.34} ± 2.86 & \textbf{67.79} ± 2.34 & \textbf{88.65} ± 0.07 & \textbf{55.67} ± 6.33            \\ \hline
\end{tabular}
\label{tab:main_result}
\end{table*}

\begin{table*}[h]
\centering
% \small
\caption{Graph Classification Results (Average ROC-AUC ± Standard Deviation) on OGBG-MOLHIV dataset. Winner in each backbone/dataset pair is highlighted in bold.}
\begin{tabular}{cccccc}
\hline
         & GraphSAGE                                                        & GCN                                                              & GIN                                                            & GAT                                                              & GXN                                                            \\ \hline
Original & 70.37   ± 0.42                           & 73.49   ± 1.99                           & 65.11 ± 2.56                           & 75.83   ± 1.78                           & 69.15 ± 0.01                           \\
CARE     & \textbf{74.33}   ± 2.12 & \textbf{74.29}   ± 1.07 & \textbf{65.42} ± 3.70 &  \textbf{76.89}   ± 2.18 & \textbf{69.17} ± 0.02 \\ \hline
\end{tabular}
\begin{tabular}{cccccc}
\hline
         & UGformer                                                        & SAGPool                                                              & DiffPool                                                            & HGPSLPool                                                              & MEWISPool                                                            \\ \hline
Original & 77.23   ± 3.54                           & 73.80   ± 1.86                           & \textbf{71.63} ± 2.25                           & 76.08   ± 2.86                           & \textbf{79.66} ± 1.71                           \\
CARE     & \textbf{78.04}   ± 3.19 & \textbf{74.42}   ± 1.53 & 70.21 ± 2.79 &  \textbf{77.23}   ± 2.16 & 77.37 ± 1.05 \\ \hline
\end{tabular}
\label{tab:ogb}
\end{table*}

\subsection{Performance Comparison with GNN Backbones}
\label{subsec:main}

\noindent
{\bf Effectiveness Analysis.} We assess the graph classification performance on the first 8 datasets and the last dataset using two different metrics. The former is assessed by the classification accuracy and the latter by the ROC-AUC. This is because the OGBG-MOLHIV dataset has a severe class imbalance issue. The results on the first 8 datasets are reported in Table \ref{tab:main_result}. Each row in the table shows the performance of an original GNN backbone and the performance after applying CARE. Each column reports the results on a dataset. In total there are 88 backbone/dataset pairs and the best result in each pair is highlighted in bold. As shown in Table \ref{tab:main_result}, CARE is a clear winner: it outperforms the GNN backbone in 84 out of 88 cases. 
CARE gains over 1\% improvement in the absolute accuracy in 57 out of 88 winning cases, while the drops in the accuracy of all losing cases are all less than 1\%. 
In particular,  the improvement of CARE is up to 11.48\%, which is achieved on the FRANKENSTEIN dataset with GraphSAGE as backbone. 
The same observation is made when testing on the OGBG-MOLHIV dataset. As shown in Table \ref{tab:ogb}, CARE outperforms the GNN backbones in most cases with improvements up to 5.63\%. To sum up, the results demonstrate that CARE is able to serve as a general framework to boost up the graph classification performance over state-of-the-art GNN models on various datasets. To match with the setting of Theorem 1, we also conduct experiments under the same parameter numbers in Appendix \ref{app:same_params}. The results demonstrate that CARE can boost up the graph classification performance without introducing additional parameters. 

\begin{figure}[h]
\centering
  \includegraphics[width=\linewidth]{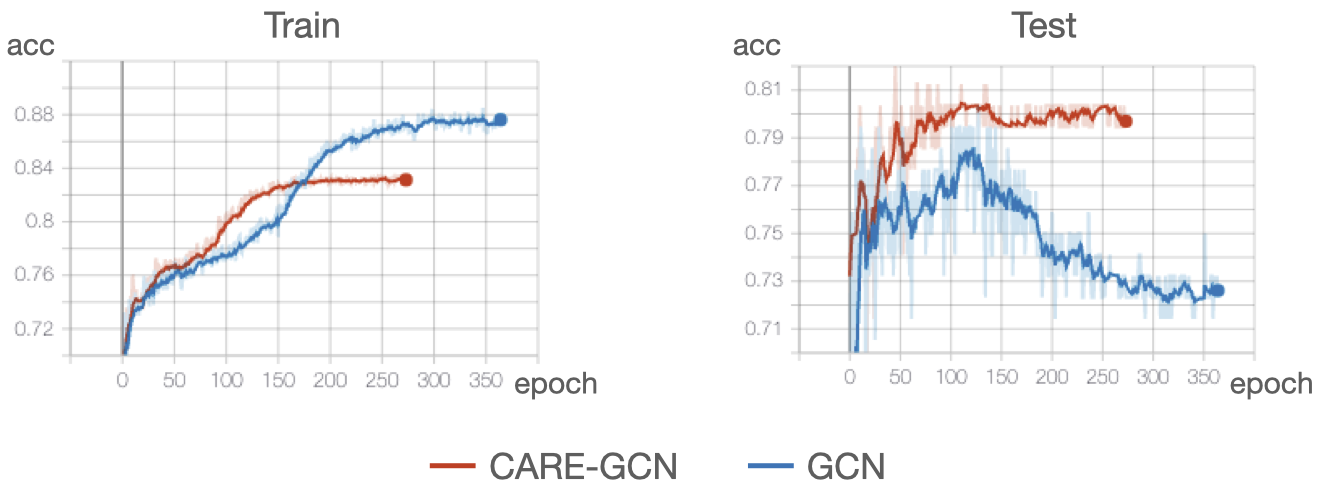}
  \caption{Accuracy curves of CARE-GCN and GCN on PROTEINS dataset.}
  \label{fig:generalization}
\end{figure}

\noindent
\textbf{Generalization Performance.} We also observe that CARE is able to alleviate the overfitting in GNN backbones. An example is shown in Figure \ref{fig:generalization}, where we plot the accuracy curves on the PROTEINS dataset with GCN as the backbone. It shows that the test accuracy of GCN (in blue) exhibits a steep and continuous downward trend starting from epoch 120, while its corresponding training accuracy continues to climb up. This indicates an obvious overfit of GCN to the training data. After applying CARE on GCN (in red), the steep drop in the test accuracy vanishes, which demonstrates the ability of CARE in remitting overfitting. 

\begin{figure}[h]
\centering
  \includegraphics[width=\linewidth]{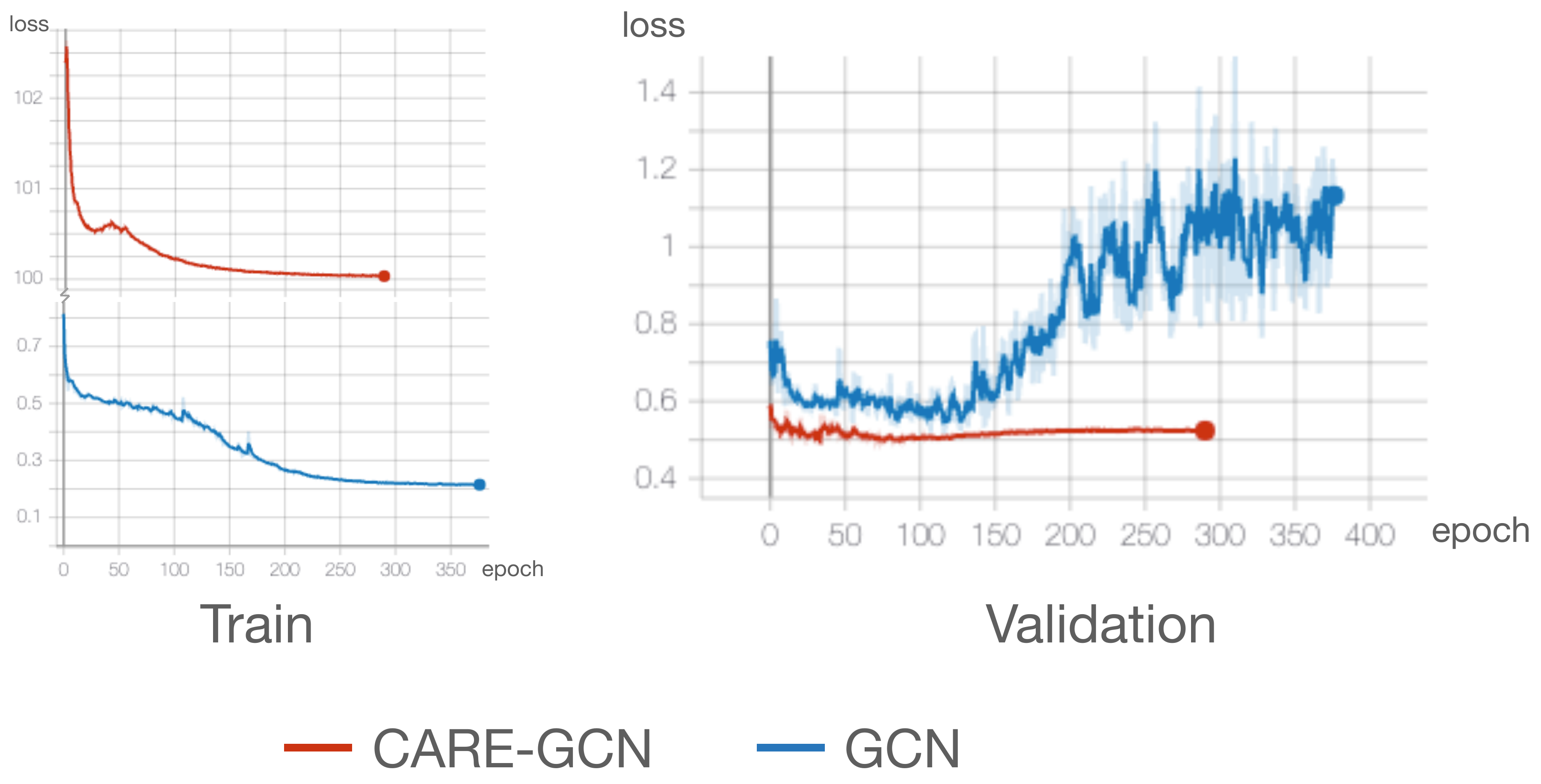}
  \caption{Loss curves of CARE-GCN and GCN on PROTEINS dataset. Note that CARE-GCN includes the class loss in the training process. The validation loss only includes the classification loss.}
  \label{fig:converge_curve}
\end{figure}

Moreover, we provide additional empirical evidence regarding the convergence behavior of the proposed method. Specifically, we present the loss curves of the GCN baseline and when it is augmented with CARE. As depicted in Figure \ref{fig:converge_curve}, while the training loss for GCN (in blue) steadily decreases throughout the training process, its validation loss begins to increase after approximately 100 epochs. In contrast, when CARE is applied alongside GCN (in red), such an increase in loss on the validation set is not observed. This illustrates that CARE not only facilitates better convergence of the model but also enables faster convergence, particularly when employing the early stopping criterion.

\subsection{Ablation Studies} 
\label{subsec:ablation}

We first conduct experiments to verify the effectiveness of our proposed Class-Aware Refiner. Then we perform two ablation studies to show how different designs of the Subgraph Selector and the loss function $\mathcal{L}$ influence the performance of CARE. \revise{We also evaluate two different choices for the similarity metric used in the class loss.}

\noindent
\textbf{Class-Aware Refiner.} To evaluate the effectiveness of the Class-Aware Refiner, we remove it from the framework and maintain the subgraph selector. To replace Eq. (3), the subgraph representation is used to refine the original graph representation:

\begin{equation}
\boldsymbol{hg}_G^{\prime} = \operatorname{Trans}([\boldsymbol{hg}_G \mid \boldsymbol{hg}^{sub}_G]),
\label{eq:refine_2}
\end{equation}

The results reported in Table \ref{tab:ablation_care} demonstrate that the classification accuracy decreases in most cases when only using the subgraph selector without Class-Aware Refiner. This observation indicates that the class representations obtained from the refiner can better enhance the graph representations.

\begin{table*}[h]
\small
\centering
\caption{Ablation Study on Class-Aware Refiner. Winner in each backbone/dataset pair is highlighted in bold.}
\begin{tabular}{cc|ccccc}
\hline
 & Refiner & D\&D           & PROTEINS       & MUTAG           & NCI1           & NCI109         \\ \hline
\multirow{2}{*}{GraphSAGE} & w/o & 72.37   ± 3.56 & 75.47   ± 3.72 & 76.58   ± 7.28 & 75.08   ± 2.19 & 73.08 ± 2.53               \\
 & w/   & \textbf{73.26}   ± 3.25 & \textbf{75.92}   ± 2.84 & \textbf{81.97}   ± 6.42  & \textbf{75.23}   ± 1.76 & \textbf{73.58}   ± 1.68 \\ \hline
\multirow{2}{*}{GCN}       & w/o & 71.53   ± 3.66 & 74.09   ± 3.91 & \textbf{80.88}   ± 7.45 & 79.29   ± 2.35 & 75.25   ± 2.51 \\
 & w/   & \textbf{72.15}   ± 3.88 & \textbf{75.01}   ± 2.91 & 79.30   ± 11.81 & \textbf{79.66}   ± 1.71 & \textbf{75.75}   ± 1.63 \\ \hline
\end{tabular}
\label{tab:ablation_care}
\end{table*}

\noindent
\textbf{Subgraph Selector.} We compare the performance of three CARE variants with different subgraph selectors on 5 datasets with 4 GNN backbones. The results are presented in Table \ref{tab:subgraph}. The first CARE variant, denoted as "None", uses the whole graph for class representation learning without selecting any subgraph. The other two models apply SAGPool and HGSPLPool respectively as the subgraph selector. The pooling ratio for both SAGPool and HGSPLPool is set to 0.5. We can see that using the subgraph selector achieves the best result in 15 out of 20 cases. When comparing the performance of using SAGPool and HGSPLPool, the former beats the latter in 14 out of 20 cases. Therefore, we choose SAGPool as the default subgraph selector.

\begin{table*}[h]
\centering
% \small
\caption{Ablation Study on Different Subgraph Selectors. Winner is highlighted in bold.}
\begin{tabular}{ccccccc}
\hline
Backbone            & Subgraph Selector & D\&D             & PROTEINS       & MUTAG           & NCI1  & NCI109         \\ \hline
\multirow{3}{*}{GraphSAGE} & None & 67.24   {± 4.64} & 75.01   {± 4.15} & \textbf{82.95}   {± 5.86} & \textbf{77.66}   {± 1.98} & 73.67 {± 1.28} \\
                     & SAGPool           & \textbf{73.26}   {± 3.25} & \textbf{75.92}   {± 2.84} & 81.97   {± 6.42}  & 75.23   {± 1.76} & 73.58 {± 1.68} \\
                     & HGPSLPool         & 71.82   {± 3.99} & 75.28   {± 3.76} & 77.57   {± 7.33}  & 75.84   {± 1.50} & \textbf{74.36} {± 2.17}\\ \hline
\multirow{3}{*}{GCN} & None              & 71.05   {± 3.89} & 73.39   {± 3.45} & 78.74   {± 9.67}  & 79.15   {± 1.66} & \textbf{76.30} {± 2.31}\\
                     & SAGPool           & \textbf{72.15}   {± 3.88} & \textbf{75.01}   {± 2.91} & \textbf{79.30}   {± 11.81} & \textbf{79.66}   {± 1.71} & 75.75 {± 1.63} \\
                     & HGPSLPool         & 68.75   {± 3.45} & 73.39   {± 3.45} & 77.66   {± 5.13}  & 79.25   {± 1.90} & 75.35 {± 2.57} \\ \hline
\multirow{3}{*}{GIN} & None              & \textbf{75.64}   {± 3.38} & 71.96   {± 5.49} & 88.80   {± 5.05}  & \textbf{82.85}   {± 1.27} & 82.11 {± 1.60} \\
                     & SAGPool           & 74.70   {± 3.37} & 72.32   {± 4.25} & 90.44   {± 4.58}  & 82.34   {± 2.11} & \textbf{82.15} {± 1.79} \\
                     & HGPSLPool         & 75.06   {± 3.49} & \textbf{72.86}   {± 4.66} & \textbf{90.47}   {± 6.10}  & 81.63   {± 1.80} & 81.05 {± 1.10}\\ \hline
\multirow{3}{*}{GAT} & None              & 74.28   {± 2.43} & 75.74   {± 2.88} & 78.22   {± 6.77}  & 75.30   {± 3.01} & 74.90 {± 2.24} \\
                     & SAGPool           & \textbf{75.38}   {± 2.93} & \textbf{76.72}   {± 1.74} & 77.69   {± 8.99}  & \textbf{78.52}   {± 2.12} & \textbf{76.39} {± 2.76} \\
                     & HGPSLPool         & 74.79   {± 2.73} & 75.46   {± 3.56} & \textbf{79.33}   {± 9.73}  & 75.74   {± 2.32} & 74.15 {± 3.53} \\ \hline
\end{tabular}
\label{tab:subgraph}
\end{table*}

\begin{table}[h]
\setlength\tabcolsep{2pt}
\centering
\small
\caption{Ablation Study on Design of Loss Function in terms of Classification Accuracy. Winner in each backbone/dataset pair is highlighted in bold.}
\begin{tabular}{@{}ccccc@{}}
\hline
Backbone  & Loss            & D\&D           & PROTEINS       & MUTAG           \\ \hline
\multirow{4}{*}{\begin{tabular}{c}Graph\\SAGE\end{tabular}} & ${\it cls}$             & 72.24   ± 2.72 & 75.82   ± 3.34 & 75.50   ± 6.05  \\
          & ${\it intra}$      & 70.12   ± 3.27 & 75.83   ± 2.83 & 77.02   ± 10.45  \\
          & ${\it inter}$       & 70.30   ± 3.61 & 75.02   ± 3.55 & 81.87   ± 7.67  \\
          & ${\it combine}$ & \textbf{73.26}   ± 3.25  & \textbf{75.92}   ± 2.84  & \textbf{81.97}   ± 6.42   \\ \hline
\multirow{4}{*}{GCN}       & ${\it cls}$             & 71.39   ± 2.81  & 74.21   ±2.74  & 77.11   ± 9.48  \\
          & ${\it intra}$      & 71.89   ± 5.35 & 74.65   ± 4.10 & 78.22   ± 7.17 \\
          & ${\it inter}$       & 71.31   ± 2.06 & 74.20   ± 3.88 & 78.18   ± 10.41 \\
          & ${\it combine}$ & \textbf{72.15}   ± 3.88  & \textbf{75.01}   ± 2.91  & \textbf{79.30} ± 11.81   \\ \hline
\end{tabular}
\label{tab:ablation_loss}
\end{table}

\noindent
\textbf{Class Loss.} We investigate different designs of the loss function to study the impact of the class loss we proposed. We consider four designs of the overall loss function $\mathcal{L}$: a) ${\it cls}$: uses the classification loss $\mathcal{L}_{cls}$ only; b) ${\it intra}$: uses a combination of the classification loss and the intra-class loss as given by $\mathcal{L} = \mathcal{L}_{cls} - \lambda_2 *  \exp(\mathcal{L}_{intra})$;
% Eq. (\ref{eq:ablation_pos}); 
c) ${\it inter}$: uses a combination of the classification loss and the inter-class loss as given by $\mathcal{L} = \mathcal{L}_{cls} + \lambda_2 * \exp(\mathcal{L}_{inter})$;
% Eq. (\ref{eq:ablation_neg}); 
and d) ${\it combine}$, the overall loss function in Eq. (\ref{eq:total_loss}). The results in Table \ref{tab:ablation_loss} show that the proposed loss function performs the best among all designs. This demonstrates the effectiveness of the proposed class loss that takes into account the intra-class similarity and inter-class discrepancy.

\noindent
\textbf{Similarity Metric for Class Loss.} We study the effect of the similarity metric in computing the class loss. Besides the cosine similarity, L2 distance is also commonly used to measure the dissimilarity between two vectors. We use L2 distance in place of the cosine similarity in Eqs. (\ref{eq:pos_loss}) and (\ref{eq:neg_loss}) to define the intra-class and inter-class losses. As L2 distance measures the dissimilarity, we take an inverse of the class loss when L2 distance is used. We then compare the performance of CARE under these two different metrics. Table \ref{tab:consistency} shows the results under GCN as the GNN backbone. It shows that CARE with the cosine similarity is more powerful than that with the L2 distance. 

\begin{equation}
\mathcal{L}_{intra} = \operatorname{norm}(\sum_{i=1}^{\mathcal{|Y|}}\operatorname{norm}(\sum_{y_G=i}\operatorname{dis}(\boldsymbol{hc}_i, \boldsymbol{hg}_G^{sub}))),
\label{eq:L2_pos}
\end{equation}

\begin{equation}
\mathcal{L}_{inter} = \operatorname{norm}(\sum_{i=1}^{\mathcal{|Y|}}\operatorname{norm}(\sum_{j=i}^{\mathcal{|Y|}}\operatorname{dis}(\boldsymbol{hc}_i, \boldsymbol{hc}_j))),
\label{eq:L2_neg}
\end{equation}

\begin{equation}
\mathcal{L}_{class} =  \exp({\mathcal{L}_{intra} - \mathcal{L}_{inter}}).
\label{eq:L2_cc_loss}
\end{equation}

\begin{table}[h]
\centering
\caption{Ablation Study on Similarity Metric for Class Loss.}
\begin{tabular}{lccc}
\hline
         & D\&D        & PROTEINS    & MUTAG       \\\hline
-        & 71.39 ± 2.81  & 74.21 ± 2.74  & 77.11 ± 9.48  \\
L2       & 71.91 ± 4.97 & 74.65 ± 3.54 & 77.69 ± 8.35 \\
KL       & 71.85 ± 4.49 & 74.38 ± 3.16 & 78.74 ± 5.19 \\
cos\_sim & \textbf{72.15} ± 3.88 & \textbf{75.01} ± 2.91  & \textbf{79.30} ± 11.81  \\\hline
\end{tabular}
\label{tab:consistency}
\end{table}

\subsection{Case Study for Class Separability} 
\label{subsec:case_study}

We design a case study to further investigate the effect of CARE, in particular its class-aware components, in refining graph representations for the classification task. The idea is to study how CARE affects the separability of graph classes in the training process. Is it able to direct the graph representation learning to move towards better class separability? In order to answer this question, we use four class separability metrics as follows. For all metrics, the larger their values, the better the class separability is. We refer the reader to Appendix \ref{app:cluster} for their formal definitions.

\begin{figure*}[h]
  \begin{center}
  \includegraphics[width=\linewidth]{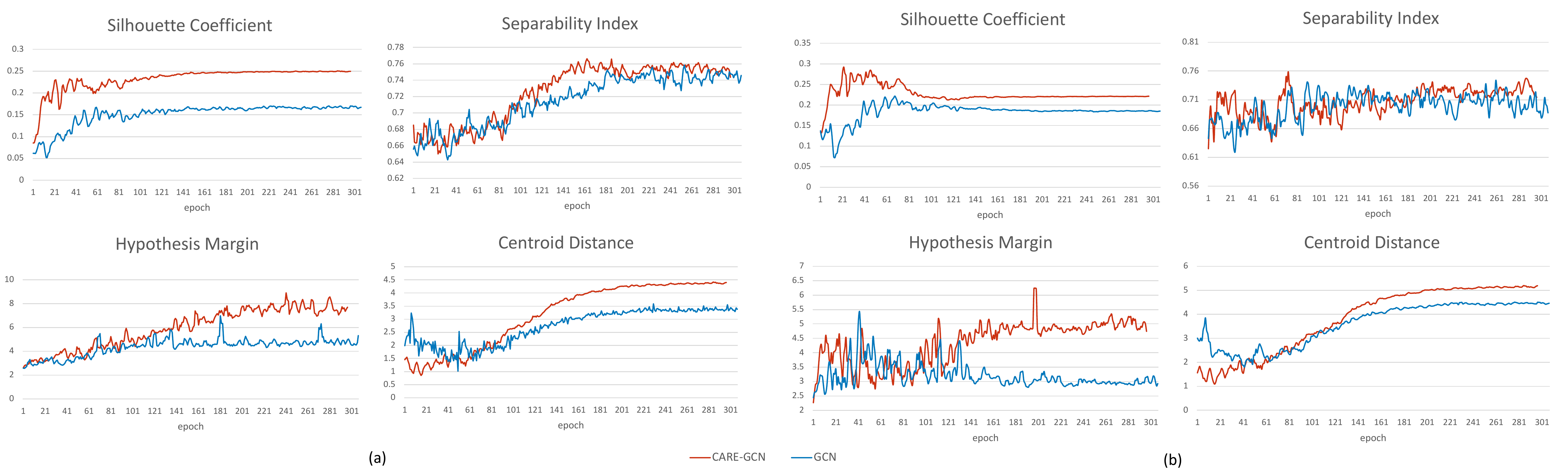}
  \end{center}
  \caption{(a) Class Separability on PROTEINS with GCN Backbone (Training Set). (b) Class Separability on PROTEINS with GCN Backbone (Test Set). The results were obtained by passing the test data once at the end of each training epoch. Note that this process doesn't affect the training in any way as the model parameters/loss are not updated when passing the test data.}
  \label{fig:case_study}
\end{figure*}

\noindent
\revise{\textbf{Silhouette Coefficient}.} It measures how similar a sample is to its own class (cohesion) compared to those from other classes (separation). Its value ranges from -1 to 1. 

\noindent
\revise{\textbf{Separability Index}.} It computes the fraction of samples that have a nearest neighbour with the same class label. Its value ranges from 0 to 1. 

\noindent
\revise{\textbf{Hypothesis Margin}.} It measures the distance between a sample’s nearest neighbor from the same class (near-hit) and the nearest neighbor from the opposing class (near-miss) and averages over all samples.

\noindent
\textbf{Centroid Distance}. It sums up the distances between the centroids for all pairs of classes, where the centroid of a class is the mean of all samples in the class. 

\begin{figure}[h]
\centering
\includegraphics[width=.8\linewidth]{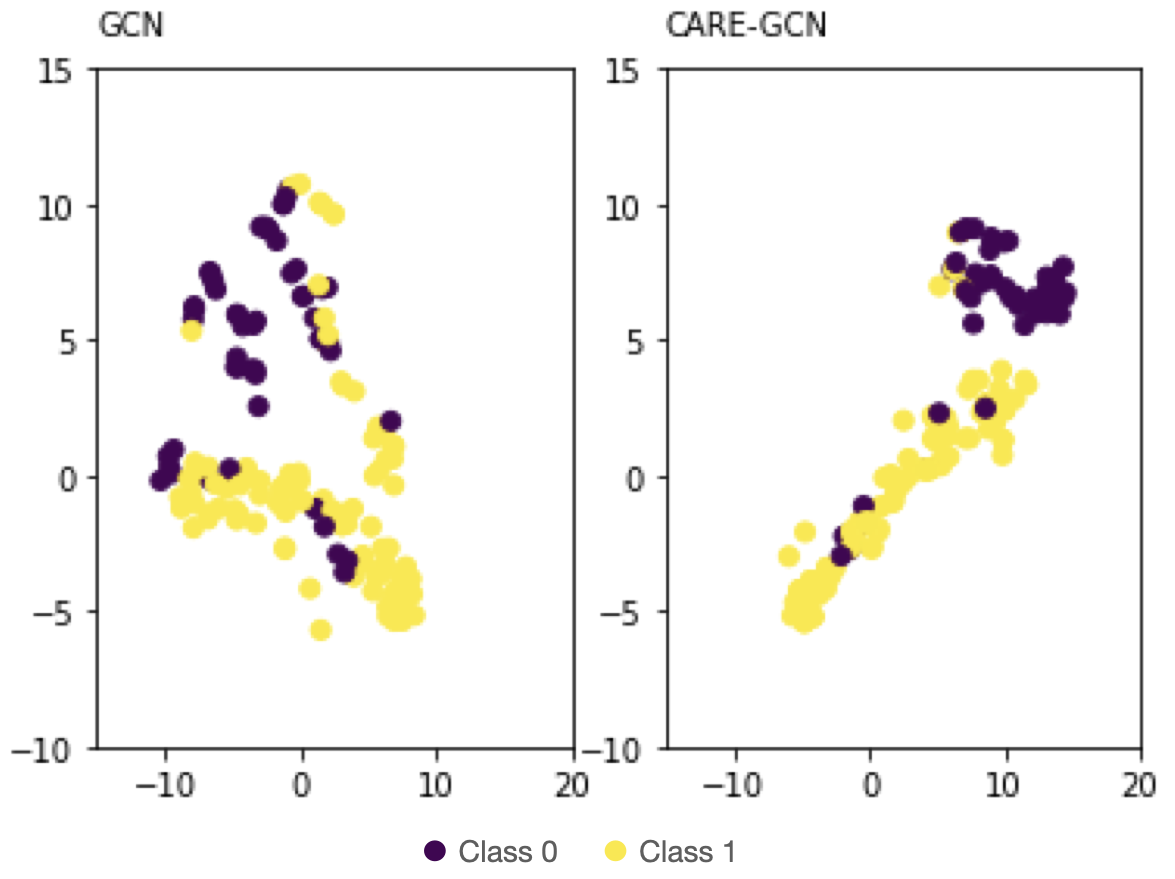}
\caption{Visualization of Graph Representations Produced by GCN and CARE-GCN on PROTEINS dataset.}
\label{fig:repr}
\end{figure}

Figure \ref{fig:case_study} reports the results on the PROTEINS dataset with GCN as the backbone. We compute the four metrics on the graph representations produced by CARE and GCN on the training data. CARE uses the refined graph representations, while GCN uses the original ones. \revise{The training curves of CARE show an upward trend across all four separability metrics. Upon convergence, CARE surpasses GCN in every metric.} In particular, CARE achieves 49.26\% improvement on Silhouette Coefficient, 1.96\% on Separability Index, 45.21\% on Hypothesis Margin, and 30.51\% on centroid distance. 
A visualization of the graph representations in each model is shown in Figure \ref{fig:repr}. Graph representations are passed into T-SNE \cite{van2008visualizing} for dimensionality reduction and colored by their class labels. This demonstrates that CARE is indeed able to steer the graph representation learning towards better class separability, which is also reflected by its superior classification performance over GNN backbones. Similar conclusions can be drawn from the results on the test data, which indicates that the class separability of CARE generalizes well to the test data. 

\subsection{Hyperparameter Analysis}
\label{subsec:hyperparameter}

In this section, we study the sensitivity of two important hyperparameters in CARE, the trade-off parameters $\lambda_1$, $\lambda_2$ in the loss function and the number of layers. We test on three datasets using GIN as backbone for this set of experiments.

\begin{table}[h]
\centering
\caption{Results when Tuning $\lambda_1$ and $\lambda_2$.}
\begin{tabular}{cc|ccc}
\hline
$\lambda_1$ & $\lambda_2$               & D\&D           & PROTEINS     & MUTAG        \\ \hline
0.1 & 0.1                   & \textbf{76.32} ± 3.33   & 72.32 ± 4.76 & 85.09 ± 6.80 \\
0.1 & 1                     & 72.76 ± 4.28   & 72.95 ± 3.81 & 85.61 ± 6.41 \\
0.1 & 10                    & 73.60 ± 2.84   & 70.61 ± 4.04 & 87.22 ± 6.00 \\
1 & 0.1                   & 74.11 ± 4.38   & 71.87 ± 14.99 & \textbf{90.47} ± 5.11 \\
1 & 1                     & 74.70 ± 3.37   & 72.32 ± 4.25 & 90.44 ± 4.58 \\
1 & 10                    & 74.45 ± 3.12   & 72.14 ± 4.71 & 89.88 ± 4.39 \\
10 & 0.1                   & 74.37 ± 3.22   & \textbf{73.14} ± 3.45 & 89.42 ± 4.64 \\
10 & 1                     & 73.85 ± 3.75   & 72.49 ± 3.84 & 89.42 ± 5.71 \\
10 & 10                    & 73.94 ± 4.40   & 70.35 ± 4.73 & 89.97 ± 5.96 \\\hline
\end{tabular}
\label{tab:lambda}
\end{table}

\noindent
\textbf{Trade-off Parameter $\lambda_1$ and $\lambda_2$}. These two hyperparameters are used in the overall loss function $\mathcal{L}$ (Eq. (\ref{eq:total_loss})) to trade-off between the class loss $\mathcal{L}_{intra}$, $\mathcal{L}_{inter}$ and the classification loss $\mathcal{L}_{cls}$. We tune the value of $\lambda_1$ and $\lambda_2$ from 0.1 to 10. The results are presented in Table \ref{tab:lambda}. It shows that the choice of $\lambda_1$ and $\lambda_2$ affects the performance marginally and there doesn't exist a value that works best for all datasets. In practice, we could use the validation set to find the best value of $\lambda_1$ and $\lambda_2$.  

\noindent
\textbf{Number of Layers}. The depth of the neural network can certainly affect the model performance. We adjust the number of layers to investigate whether CARE can adapt to different depths of neural networks. We vary the number of layers from 2 to 5, and report the results in Table \ref{tab:layer_num}. For each dataset, we underline the best result among all the numbers of layers tested. As shown in Table \ref{tab:layer_num}, CARE consistently outperforms GIN at different number of layers, except for 4 layers on PROTEINS where the performance difference is marginal at 0.09\%. The best results are achieved with 2 layers on D\&D and PROTEINS and with 5 layers on MUTAG. Therefore, the number of layers should also be selected through the validation process for different datasets.

\begin{table}[h]
\centering
\caption{Results when Tuning Number of Layers.}
\begin{tabular}{ccccc}
\hline
Layer\# & module & \multicolumn{1}{c}{D\&D} & \multicolumn{1}{c}{PROTEINS} & \multicolumn{1}{c}{MUTAG} \\ \hline
\multirow{2}{*}{2} & GIN    & 74.11 ± 3.42 & 72.42 ± 2.06   & 89.91 ± 4.35 \\
                  & CARE & \underline{\textbf{76.40} ± 2.14} & \underline{\textbf{73.22} ± 2.78}   & \textbf{90.47} ± 5.11 \\ \hline
\multirow{2}{*}{3} & GIN    & 74.53 ± 3.36 & 70.79 ± 5.18   & 87.78 ± 4.07 \\
                  & CARE & \textbf{75.13} ± 3.39 & \textbf{71.69} ± 4.65   & \textbf{90.47} ± 5.11 \\ \hline
\multirow{2}{*}{4} & GIN    & 73.10 ± 2.44 & \textbf{72.41} ± 4.45   & 89.36 ± 4.71 \\
                  & CARE & \textbf{74.70} ± 3.37 & 72.32 ± 4.25    & \textbf{90.44} ± 4.58 \\ \hline
\multirow{2}{*}{5} & GIN    & 73.93 ± 2.62 & 70.71   ± 4.00 & 91.49 ± 4.83 \\
                  & CARE & \textbf{74.70} ± 3.50 & \textbf{72.69} ± 3.24   & \underline{\textbf{91.52} ± 5.39} \\ \hline
\end{tabular}
\label{tab:layer_num}
\end{table}

\revise{From the two hyperparameter analyses above, it is clear that while CARE can enhance the performance of various GNNs, it still requires hyperparameter tuning to identify the optimal configuration.} This reliance on hyperparameter selection could pose a limitation to the practical application of CARE. In the future, further exploration is warranted to mitigate its sensitivity to hyperparameters, thereby enhancing the versatility of CARE.

\begin{table*}[h]
\centering
\caption{Time Efficiency of CARE and Backbones. Total time (h) was recorded for a single run (including training, validation, and test) with batch size 20 and 10-fold CV. Best time in each backbone/dataset pair is highlighted in bold.}
\begin{tabular}{cc|cc|cc|cc}
\hline
\multicolumn{2}{c|}{\multirow{2}{*}{Model}} & \multicolumn{2}{c|}{D\&D}              & \multicolumn{2}{c|}{PROTEINS}       & \multicolumn{2}{c}{MUTAG}           \\
\multicolumn{2}{c|}{}                       & Epoch \#   ± s.d. &  Time & Epoch \#   ± s.d. &  Time & Epoch \#   ± s.d. &  Time \\ \hline
\multirow{2}{*}{GraphSAGE}    & Original    & 500.6 {± 123.2}     & 1.205           & 320.5 {± 53.2}      & 1.209          & 384.1    {± 101.3}  & 0.180          \\
                     & CARE     & 293.5 {± 12.0} & \textbf{1.142} & 282.0 {± 51.5} & \textbf{0.911} & 302.2 {± 34.1} & \textbf{0.159} \\ \hline
\multirow{2}{*}{GCN} & Original & 267.4 {± 3.4}  & \textbf{0.692} & 365.0 {± 27.9} & 0.670 & 352.4 {± 69.9} & \textbf{0.132} \\
                     & CARE     & 264.1 {± 5.2}  & 0.848 & 306.5 {± 17.2} & \textbf{0.665} & 332.4 {± 57.3} & 0.143 \\ \hline
\end{tabular}
\label{tab:time}
\end{table*}

\subsection{Time Efficiency}
\label{subsec:time_efficiency}
CARE, when applied to a GNN backbone, introduces an additional refiner and the class loss. A natural question arises: will CARE significantly sacrifice the efficiency of its GNN backbone for better classification performance? This subsection aims to answer this question. Table \ref{tab:time} reports the number of epochs and the total time needed (including training, validation and test) for CARE and the backbones GraphSAGE and GCN. It can be seen that CARE takes less number of epochs to converge than its GNN counterpart in all cases. Consequently, the running time of CARE is shorter than (4 out of 6 cases) or comparable to its backbones. The results demonstrate that CARE is able to work on top of existing GNN models with superior effectiveness and improved/comparable efficiency, making it a practical choice in real applications. The high efficiency observed can be attributed to the design of class representations. As the class representation is updated based on (sub)graph representations, it introduces only linear time complexity as the number of training samples increases. Consequently, this design enables the method to be applied to various graph data, irrespective of the size of the graphs, thereby unleashing its potential for adaptability.

\section{Discussion}
\label{subsec:discussion}

Our proposed CARE addresses two significant limitations GNNs in graph classification tasks: neglect of graph-level relationships and generalization issues. By seamlessly integrating CARE into GNN models, we effectively incorporate graph-level relationships, leading to performance improvements and enhanced generalization ability without augmenting the time and memory complexity of the base model.

While CARE presents a promising approach for improving graph classification performance, it is essential to acknowledge its limitations and potential avenues for future research:

\begin{itemize}
\item \textbf{Hyperparameter Sensitivity.} One limitation of CARE is its sensitivity to hyperparameters, which may require manual tuning for optimal performance. Future work could focus on developing strategies to reduce this sensitivity, thereby enhancing the versatility and ease of application of CARE across different datasets and tasks.
\item \textbf{Theoretical Understanding.} Although we provide theoretical support for CARE's generalization properties, further theoretical analysis of convergence behavior could provide valuable insights into the underlying mechanisms of CARE.
\end{itemize}

\section{Conclusions}

In this paper, we proposed CARE, a novel graph representation refinement framework for GNN-based graph classification. It fills the gap that existing GNNs fail to consider graph-level relationships in model design, and meanwhile improves the model generalization as proven theoretically and evidenced empirically. Its plug-in-play nature makes it a powerful framework to build upon arbitrary GNN models and boost up their classification performance without sacrificing efficiency. In the future, we will further explore the convergence behavior of such plugin component.

\section*{Acknowledgment}
\revise{This research/project is supported by the National Research Foundation, Singapore under its Industry Alignment Fund – Pre-positioning (IAF-PP) Funding Initiative, and the Ministry of Education, Singapore under its MOE Academic Research Fund Tier 2 (STEM RIE2025 Award MOE-T2EP20220-0006).} Any opinions, findings and conclusions or recommendations expressed in this material are those of the author(s) and do not reflect the views of National Research Foundation, Singapore, and the Ministry of Education, Singapore.

\appendix
\section{Theoretical Proofs}
\label{app:proofs}

\subsection{Proof Sketch of Lemma 1}
\label{app:lemma}
Our proof follows the same flow as Lemma 1 in  \cite{kabkab2016size}.

A parametrized class of functions with parameters in $\mathbb{R}^t$ that is computable in no more than $p$ operations has a VC dimension which is $O(t^2 p^2)$ \cite{bartlett2003vapnik}. $t$ in GCN and CARE can be formulated as: 

\begin{equation}
\label{eq:tgcn}
t_{GCN} = \sum_{l=0}^d h_{gcn_{in}}^l h_{gcn_{out}}^l; 
\end{equation}

\begin{equation}
\label{eq:tcare}
\small
\begin{split}
t_{CARE} = \sum_{l=0}^d (h_{gcn_{in}}^l h_{gcn_{out}}^l + h_{gcn_{out}}^l + h_{set_{in}}^l h_{set_{out}}^l \\ + h_{trans_{in}}^l h_{trans_{out}}^l). 
\end{split}
\end{equation}

By plugging in the number of multiplications $q_1(d)$ and $q_2(d)$ given by Eqs. (\ref{eq:q1}) and (\ref{eq:q2}), together with the above equations on the number of parameters $t$, into $O(t^2p^2)$, we complete the proof of Lemma 1 for both GCN and CARE.

\subsection{Proof of Theorem 1}
\label{app:vcdim}

We compare the VC dimension upper bounds of a GCN layer and a GCN-based CARE layer under the identical number of parameters. According to Section 2.2 of \cite{abu2012learning}, the VC dimension provides a loose generalization bound for models and can be used as a guideline for generalization comparison - models with a lower upper bound tend to have better generalization capability. The number of parameters in a GCN layer $t_1$ and that in a GCN-based CARE layer $t_2$ are formulated as: 

\begin{equation}
    t_1 = h_{gcn_{in}} h_{gcn_{out}},
\end{equation}

\begin{equation}
\begin{split}
    t_2 = h_{gcn_{in}} h_{gcn_{out}} + h_{gcn_{out}} + h_{set_{in}} h_{set_{out}} \\ + h_{trans_{in}} h_{trans_{out}}.
\end{split}
\end{equation}

In our setting, we choose a basic hidden dimension $h_1$ and $h_2$ for GCN and CARE respectively. We set each layer to be an integer multiple of the basic hidden dimension. Thus, $t_1 = h_1^2$ and $t_2 = h_2^2 + h_2 + h_2^2 + 2 h_2^2 = 4 h_2^2 + h_2$, respectively.

Note that $h_{trans_{in}} = h_{set_{out}} + h_{gcn_{out}}$ as we concatenate the class representation with the subgraph representation.

Similarly, the computational complexities $q_1$ and $q_2$ can be rewritten as:

\begin{equation}
    q_1(d) = \sum_{l=0}^d (n h_1^2 + n^2 h_1),
\end{equation}

\begin{equation}
\begin{split}
    q_2(d) &= \sum_{l=0}^d (4nh_2^2 + (2n^2 + n)h_2).
\end{split}
\end{equation}

When $d = 1$, the complexity can be written as:

\begin{equation}
    q_1(1) = n h_1^2 + n^2 h_1,
\end{equation}

\begin{equation}
\begin{split}
    q_2(1) &= 4nh_2^2 + (2n^2 + n)h_2.
\end{split}
\end{equation}

Under the identical number of parameters, we let $t_1 = t_2$, and have $h_1 = \sqrt{4h_2^2 + h_2}$. Thus, 

\begin{equation}
    q_1(1) = 4nh_2^2 + nh_2 + n^2\sqrt{4h_2^2 + h_2}.
\end{equation}

The difference between $q_1(1)$ and $q_2(1)$ satisfies:

\begin{equation}
\label{eq:q_minus1}
\begin{split}
    q_1(1) - q_2(1) = n^2(\sqrt{4h_2^2 + h_2} - 2h_2).
\end{split}
\end{equation}

Because $\sqrt{4h_2^2 + h_2} - 2h_2 > 0$, we have:

\begin{equation}
\label{eq:conclusion1}
    q_1(1) > q_2(1).
\end{equation}

According to our setting, the input and output feature map sizes of all layers is identical, which means that the `\textit{n}' in each layer's complexity equation are identical. Thus, we extend Eq. (\ref{eq:conclusion1}) to the full model and have:

\begin{equation}
\label{eq:conclusion2}
    q_1(d) > q_2(d).
\end{equation}

With Eq. (\ref{eq:conclusion2}) and Lemma 1, we complete the proof of Theorem 1.

\section{Effectiveness Analysis under the same Parameter Number}
\label{app:same_params}

CARE is proposed as a plug-and-play framework. However, in addition to directly plugging it in a GNN backbone without changing the number of model parameters in the backbone (as what we have done in experiments in the submitted version), it could also be used in a way that the resultant CARE after plug-in has a comparable number of parameters to the original GNN backbone before plug-in. This can be achieved by adjusting the number of parameters in the GNN backbone at the time when CARE is plugged in. To demonstrate this, we conduct a new experiments to match with the setting of Theorem 1. For each GNN backbone, we first set its number of parameters to 100K. For CARE, we adjust the hidden dimension of each GNN backbone to which CARE is applied such that the number of parameters of CARE is also 100K.  As shown in Table \ref{tab:same_params_exp}, CARE still outperforms its GNN backbone in 8 out of 9 cases. The results demonstrate that CARE can boost up the graph classification performance without introducing additional parameters.

\begin{table}[h]
\centering
\caption{Graph Classification Results (Average Accuracy ± Standard Deviation) under the same parameters setting. The parameter numbers of all models are 100K. Winner in each backbone/dataset pair is highlighted in bold.}
\begin{tabular}{lc|ccc}
\hline
                           &          & DD           & PROTEINS     & MUTAG           \\ \hline
\multirow{2}{*}{GraphSAGE} & original & 72.18 ± 2.93 & 74.87 ± 3.38 & 75.48    ± 6.11 \\
                           & CARE     & \textbf{72.22} ± 3.10 & \textbf{75.74} ± 1.68 & \textbf{76.08} ± 10.83   \\ \hline
\multirow{2}{*}{GCN}       & original & 71.02 ± 3.17 & 73.89 ± 2.85 & 77.52 ± 10.81   \\
                           & CARE     & \textbf{71.73} ± 4.12 & \textbf{74.91} ± 3.59 & \textbf{79.27} ± 4.31    \\ \hline
\multirow{2}{*}{GIN}       & original & 73.10 ± 2.44 & 72.41 ± 4.45 & 89.36 ± 4.71    \\
                           & CARE     & \textbf{73.19} ± 4.44 & \textbf{70.43} ± 4.69 & \textbf{89.70} ± 5.53    \\ \hline
\end{tabular}
\label{tab:same_params_exp}
\end{table}

\section{Class Separability Metrics in Case Study}
\label{app:cluster}

\subsection{Silhouette Coefficient}
\label{app:silhouette}

The Silhouette of a sample $x_i$ is defined as:

\begin{equation}
    sil(x_i) = \frac{b^i-a^i}{max(a^i, b^i)},
\end{equation}

\begin{equation}
    Silhouette = \underset{x_i}{AVERAGE}(\{sil(x_i)\}),
\end{equation}

\noindent
where $a^i$ denotes the average distance between $x_i$ and all other samples in the same class, and $b^i$ denotes the smallest mean distance from $x_i$ to all samples in any other class.  

\subsection{Separability Index}
\label{app:si}

The Separability Index $SI$ is defined as:

\begin{equation}
    K(x_i, x_j) = 
    \begin{cases}
    1, & \text{if } y_i = y_j, y_i \in \mathcal{Y}, y_j \in \mathcal{Y}\\
    0, & \text{otherwise}
    \end{cases}
    ,
\end{equation}

\begin{equation}
    x^{\prime}_i = \underset{x_j \neq x_i}{\arg\min}(\|x_i - x_j\|),
\end{equation}

\begin{equation}
    SI = \frac{\sum_{x_i} K(x_i, x_i^{\prime})}{m},
\end{equation}

\noindent
where $m$ is the total number of samples, $x_i$ denotes the $i$-th sample, $y_i$ denotes its corresponding class label, and  $\mathcal{Y}$ denotes the set of classes. The nearest neighbour distance function $\|\cdot\|$ is assumed to utilise a suitable metric, e.g., a Manhalobis metric for symbolic data or a Euclidean metric for spatial data.

\subsection{Hypothesis Margin}
\label{app:hm}

The Hypothesis Margin ($HM$) is defined as:

\begin{equation}
    hm(x_i) = \frac{\|x_i-\textbf{nearmiss}(x_i)\|}{\|x_i-\textbf{nearhit}(x_i)\|},
\end{equation}

\begin{equation}
    HM = \underset{x_i}{AVERAGE}(\{hm(x_i)\}),
\end{equation}

\noindent
where \textbf{nearhit}($x_i$) and \textbf{nearmiss}($x_i$) denote the nearest sample to $x_i$ with the same and different label, respectively. $\|\cdot\|$ denotes the L2 distance.  Note that a chosen set of features affects the margin through the distance measure.

\subsection{Centroid Distance}
\label{app:dis}

The Centroid Distance ($CD$) is defined as:

\begin{equation}
    c_i = AVERAGE(\{x\}_{y_x=i}),
\end{equation}

\begin{equation}
CD = \sum_{i=1}^{\mathcal{|Y|}}\sum_{j=i}^{\mathcal{|Y|}} \|c_i - c_j\|,
\end{equation}

\noindent
where $y_x$ denotes the class label of a sample $x$, $\mathcal{Y}$ denotes the set of classes and $\|\cdot\|$ denotes the L2 distance. 

\ifCLASSOPTIONcaptionsoff
  \newpage
\fi

\bibliographystyle{IEEEtran}
\bibliography{ref}

\begin{thebibliography}{10}
\providecommand{\url}[1]{#1}
\csname url@samestyle\endcsname
\providecommand{\newblock}{\relax}
\providecommand{\bibinfo}[2]{#2}
\providecommand{\BIBentrySTDinterwordspacing}{\spaceskip=0pt\relax}
\providecommand{\BIBentryALTinterwordstretchfactor}{4}
\providecommand{\BIBentryALTinterwordspacing}{\spaceskip=\fontdimen2\font plus
\BIBentryALTinterwordstretchfactor\fontdimen3\font minus
  \fontdimen4\font\relax}
\providecommand{\BIBforeignlanguage}[2]{{%
\expandafter\ifx\csname l@#1\endcsname\relax
\typeout{** WARNING: IEEEtran.bst: No hyphenation pattern has been}%
\typeout{** loaded for the language `#1'. Using the pattern for}%
\typeout{** the default language instead.}%
\else
\language=\csname l@#1\endcsname
\fi
#2}}
\providecommand{\BIBdecl}{\relax}
\BIBdecl

\bibitem{xu2018powerful}
K.~Xu, W.~Hu, J.~Leskovec, and S.~Jegelka, ``How powerful are graph neural
  networks?'' in \emph{International Conference on Learning Representations},
  2018.

\bibitem{ying2018hierarchical}
R.~Ying, J.~You, C.~Morris, X.~Ren, W.~L. Hamilton, and J.~Leskovec,
  ``Hierarchical graph representation learning with differentiable pooling,''
  \emph{arXiv preprint arXiv:1806.08804}, 2018.

\bibitem{li2023imf}
X.~Li, X.~Zhao, J.~Xu, Y.~Zhang, and C.~Xing, ``Imf: interactive multimodal
  fusion model for link prediction,'' in \emph{Proceedings of the ACM Web
  Conference 2023}, 2023, pp. 2572--2580.

\bibitem{kipf2016variational}
T.~N. Kipf and M.~Welling, ``Variational graph auto-encoders,'' \emph{arXiv
  preprint arXiv:1611.07308}, 2016.

\bibitem{kipf2016semi}
------, ``Semi-supervised classification with graph convolutional networks,''
  \emph{arXiv preprint arXiv:1609.02907}, 2016.

\bibitem{hamilton2017inductive}
W.~L. Hamilton, R.~Ying, and J.~Leskovec, ``Inductive representation learning
  on large graphs,'' in \emph{Proceedings of the 31st International Conference
  on Neural Information Processing Systems}, 2017, pp. 1025--1035.

\bibitem{gilmer2017neural}
J.~Gilmer, S.~S. Schoenholz, P.~F. Riley, O.~Vinyals, and G.~E. Dahl, ``Neural
  message passing for quantum chemistry,'' in \emph{International conference on
  machine learning}.\hskip 1em plus 0.5em minus 0.4em\relax PMLR, 2017, pp.
  1263--1272.

\bibitem{bian2023cpmr}
Q.~Bian, J.~Xu, H.~Fang, and Y.~Ke, ``Cpmr: Context-aware incremental
  sequential recommendation with pseudo-multi-task learning,'' in
  \emph{Proceedings of the 32nd ACM International Conference on Information and
  Knowledge Management}, 2023, pp. 120--130.

\bibitem{10508252}
J.~Xu, Q.~Bian, X.~Li, A.~Zhang, Y.~Ke, M.~Qiao, W.~Zhang, W.~K.~J. Sim, and
  B.~Gulyás, ``Contrastive graph pooling for explainable classification of
  brain networks,'' \emph{IEEE Transactions on Medical Imaging}, pp. 1--1,
  2024.

\bibitem{lee2019self}
J.~Lee, I.~Lee, and J.~Kang, ``Self-attention graph pooling,'' in
  \emph{International Conference on Machine Learning}.\hskip 1em plus 0.5em
  minus 0.4em\relax PMLR, 2019, pp. 3734--3743.

\bibitem{wu2020comprehensive}
Z.~Wu, S.~Pan, F.~Chen, G.~Long, C.~Zhang, and S.~Y. Philip, ``A comprehensive
  survey on graph neural networks,'' \emph{IEEE transactions on neural networks
  and learning systems}, vol.~32, no.~1, pp. 4--24, 2020.

\bibitem{zhang2019hierarchical}
Z.~Zhang, J.~Bu, M.~Ester, J.~Zhang, C.~Yao, Z.~Yu, and C.~Wang, ``Hierarchical
  graph pooling with structure learning,'' \emph{arXiv preprint
  arXiv:1911.05954}, 2019.

\bibitem{nouranizadeh2021maximum}
A.~Nouranizadeh, M.~Matinkia, M.~Rahmati, and R.~Safabakhsh, ``Maximum entropy
  weighted independent set pooling for graph neural networks,'' \emph{arXiv
  preprint arXiv:2107.01410}, 2021.

\bibitem{song2021network}
X.~Song, R.~Ma, J.~Li, M.~Zhang, and D.~P. Wipf, ``Network in graph neural
  network,'' \emph{arXiv preprint arXiv:2111.11638}, 2021.

\bibitem{papp2021dropgnn}
P.~A. Papp, K.~Martinkus, L.~Faber, and R.~Wattenhofer, ``Dropgnn: random
  dropouts increase the expressiveness of graph neural networks,''
  \emph{Advances in Neural Information Processing Systems}, vol.~34, 2021.

\bibitem{ding2018semi}
M.~Ding, J.~Tang, and J.~Zhang, ``Semi-supervised learning on graphs with
  generative adversarial nets,'' in \emph{Proceedings of the 27th ACM
  International Conference on Information and Knowledge Management}, 2018, pp.
  913--922.

\bibitem{you2020graph}
Y.~You, T.~Chen, Y.~Sui, T.~Chen, Z.~Wang, and Y.~Shen, ``Graph contrastive
  learning with augmentations,'' \emph{Advances in Neural Information
  Processing Systems}, vol.~33, pp. 5812--5823, 2020.

\bibitem{byrd2019effect}
J.~Byrd and Z.~Lipton, ``What is the effect of importance weighting in deep
  learning?'' in \emph{International Conference on Machine Learning}.\hskip 1em
  plus 0.5em minus 0.4em\relax PMLR, 2019, pp. 872--881.

\bibitem{lin2017focal}
T.-Y. Lin, P.~Goyal, R.~Girshick, K.~He, and P.~Doll{\'a}r, ``Focal loss for
  dense object detection,'' in \emph{Proceedings of the IEEE international
  conference on computer vision}, 2017, pp. 2980--2988.

\bibitem{wen2016discriminative}
Y.~Wen, K.~Zhang, Z.~Li, and Y.~Qiao, ``A discriminative feature learning
  approach for deep face recognition,'' in \emph{European conference on
  computer vision}.\hskip 1em plus 0.5em minus 0.4em\relax Springer, 2016, pp.
  499--515.

\bibitem{vapnik2015uniform}
V.~N. Vapnik and A.~Y. Chervonenkis, ``On the uniform convergence of relative
  frequencies of events to their probabilities,'' in \emph{Measures of
  complexity}.\hskip 1em plus 0.5em minus 0.4em\relax Springer, 2015, pp.
  11--30.

\bibitem{velivckovic2017graph}
P.~Veli{\v{c}}kovi{\'c}, G.~Cucurull, A.~Casanova, A.~Romero, P.~Lio, and
  Y.~Bengio, ``Graph attention networks,'' \emph{arXiv preprint
  arXiv:1710.10903}, 2017.

\bibitem{weisfeiler1968reduction}
B.~Weisfeiler and A.~Leman, ``The reduction of a graph to canonical form and
  the algebra which appears therein,'' \emph{NTI, Series}, vol.~2, no.~9, pp.
  12--16, 1968.

\bibitem{vaswani2017attention}
A.~Vaswani, N.~Shazeer, N.~Parmar, J.~Uszkoreit, L.~Jones, A.~N. Gomez,
  {\L}.~Kaiser, and I.~Polosukhin, ``Attention is all you need,'' in
  \emph{Advances in neural information processing systems}, 2017, pp.
  5998--6008.

\bibitem{nguyen2019universal}
D.~Q. Nguyen, T.~D. Nguyen, and D.~Phung, ``Universal graph transformer
  self-attention networks,'' \emph{arXiv preprint arXiv:1909.11855}, 2019.

\bibitem{rong2020grover}
Y.~Rong, Y.~Bian, T.~Xu, W.~Xie, Y.~Wei, W.~Huang, and J.~Huang, ``Grover:
  Self-supervised message passing transformer on large-scale molecular data,''
  \emph{Advances in Neural Information Processing Systems}, 2020.

\bibitem{yu2020graph}
J.~Yu, T.~Xu, Y.~Rong, Y.~Bian, J.~Huang, and R.~He, ``Graph information
  bottleneck for subgraph recognition,'' \emph{arXiv preprint
  arXiv:2010.05563}, 2020.

\bibitem{huang2022graph}
C.~Huang, M.~Li, F.~Cao, H.~Fujita, Z.~Li, and X.~Wu, ``Are graph convolutional
  networks with random weights feasible?'' \emph{IEEE Transactions on Pattern
  Analysis and Machine Intelligence}, vol.~45, no.~3, pp. 2751--2768, 2022.

\bibitem{zhang2018end}
M.~Zhang, Z.~Cui, M.~Neumann, and Y.~Chen, ``An end-to-end deep learning
  architecture for graph classification,'' in \emph{Thirty-Second AAAI
  Conference on Artificial Intelligence}, 2018.

\bibitem{baek2021accurate}
J.~Baek, M.~Kang, and S.~J. Hwang, ``Accurate learning of graph representations
  with graph multiset pooling,'' \emph{arXiv preprint arXiv:2102.11533}, 2021.

\bibitem{yang2021soft}
M.~Yang, Y.~Shen, H.~Qi, and B.~Yin, ``Soft-mask: Adaptive substructure
  extractions for graph neural networks,'' in \emph{Proceedings of the Web
  Conference 2021}, 2021, pp. 2058--2068.

\bibitem{rong2019dropedge}
Y.~Rong, W.~Huang, T.~Xu, and J.~Huang, ``Dropedge: Towards deep graph
  convolutional networks on node classification,'' \emph{arXiv preprint
  arXiv:1907.10903}, 2019.

\bibitem{feng2019graph}
F.~Feng, X.~He, J.~Tang, and T.-S. Chua, ``Graph adversarial training:
  Dynamically regularizing based on graph structure,'' \emph{IEEE Transactions
  on Knowledge and Data Engineering}, vol.~33, no.~6, pp. 2493--2504, 2019.

\bibitem{thakoor2021bootstrapped}
S.~Thakoor, C.~Tallec, M.~G. Azar, R.~Munos, P.~Veli{\v{c}}kovi{\'c}, and
  M.~Valko, ``Bootstrapped representation learning on graphs,'' \emph{arXiv
  preprint arXiv:2102.06514}, 2021.

\bibitem{sun2021sugar}
Q.~Sun, J.~Li, H.~Peng, J.~Wu, Y.~Ning, P.~S. Yu, and L.~He, ``Sugar: Subgraph
  neural network with reinforcement pooling and self-supervised mutual
  information mechanism,'' in \emph{Proceedings of the Web Conference 2021},
  2021, pp. 2081--2091.

\bibitem{zhou2020k}
Q.~Zhou, B.~Sun, Y.~Song, and S.~Li, ``K-means clustering based undersampling
  for lower back pain data,'' in \emph{Proceedings of the 2020 3rd
  International Conference on Big Data Technologies}, 2020, pp. 53--57.

\bibitem{shi2021boosting}
S.~Shi, K.~Qiao, S.~Yang, L.~Wang, J.~Chen, and B.~Yan, ``Boosting-gnn:
  Boosting algorithm for graph networks on imbalanced node classification,''
  \emph{Frontiers in neurorobotics}, p. 154, 2021.

\bibitem{wu2023rornet}
Y.~Wu, Y.~Zhang, W.~Ma, M.~Gong, X.~Fan, M.~Zhang, A.~Qin, and Q.~Miao,
  ``Rornet: Partial-to-partial registration network with reliable overlapping
  representations,'' \emph{IEEE Transactions on Neural Networks and Learning
  Systems}, 2023.

\bibitem{zaheer2017deep}
M.~Zaheer, S.~Kottur, S.~Ravanbakhsh, B.~Poczos, R.~R. Salakhutdinov, and A.~J.
  Smola, ``Deep sets,'' \emph{Advances in neural information processing
  systems}, vol.~30, 2017.

\bibitem{CharlesRQi2016PointNetDL}
C.~R. Qi, H.~Su, K.~Mo, and L.~J. Guibas, ``Pointnet: Deep learning on point
  sets for 3d classification and segmentation,'' \emph{computer vision and
  pattern recognition}, 2016.

\bibitem{cox1958regression}
D.~R. Cox, ``The regression analysis of binary sequences,'' \emph{Journal of
  the Royal Statistical Society: Series B (Methodological)}, vol.~20, no.~2,
  pp. 215--232, 1958.

\bibitem{vapnik2000nature}
V.~Vapnik, ``The nature of statistical learning theory (information science and
  statistics) springer-verlag,'' \emph{New York}, 2000.

\bibitem{bartlett2003vapnik}
P.~L. Bartlett and W.~Maass, ``Vapnik-chervonenkis dimension of neural nets,''
  \emph{The handbook of brain theory and neural networks}, pp. 1188--1192,
  2003.

\bibitem{kabkab2016size}
M.~Kabkab, E.~Hand, and R.~Chellappa, ``On the size of convolutional neural
  networks and generalization performance,'' in \emph{2016 23rd International
  Conference on Pattern Recognition (ICPR)}.\hskip 1em plus 0.5em minus
  0.4em\relax IEEE, 2016, pp. 3572--3577.

\bibitem{KKMMN2016}
\BIBentryALTinterwordspacing
K.~Kersting, N.~M. Kriege, C.~Morris, P.~Mutzel, and M.~Neumann, ``Benchmark
  data sets for graph kernels,'' 2016. [Online]. Available:
  \url{http://graphkernels.cs.tu-dortmund.de}
\BIBentrySTDinterwordspacing

\bibitem{hu2020open}
W.~Hu, M.~Fey, M.~Zitnik, Y.~Dong, H.~Ren, B.~Liu, M.~Catasta, and J.~Leskovec,
  ``Open graph benchmark: Datasets for machine learning on graphs,''
  \emph{Advances in neural information processing systems}, vol.~33, pp.
  22\,118--22\,133, 2020.

\bibitem{wijesinghe2021new}
A.~Wijesinghe and Q.~Wang, ``A new perspective on" how graph neural networks go
  beyond weisfeiler-lehman?",'' in \emph{International Conference on Learning
  Representations}, 2021.

\bibitem{li2020graph}
M.~Li, S.~Chen, Y.~Zhang, and I.~W. Tsang, ``Graph cross networks with vertex
  infomax pooling,'' \emph{arXiv preprint arXiv:2010.01804}, 2020.

\bibitem{van2008visualizing}
L.~Van~der Maaten and G.~Hinton, ``Visualizing data using t-sne.''
  \emph{Journal of machine learning research}, vol.~9, no.~11, 2008.

\bibitem{abu2012learning}
Y.~S. Abu-Mostafa, M.~Magdon-Ismail, and H.-T. Lin, \emph{Learning from
  data}.\hskip 1em plus 0.5em minus 0.4em\relax AMLBook New York, 2012, vol.~4.

\end{thebibliography}
\end{document}